\documentclass[11pt]{article}
\usepackage[utf8]{inputenc}
\usepackage[T1]{fontenc}
\usepackage{lmodern}
\usepackage[a4paper,margin=1in]{geometry}
\usepackage{amsmath,amssymb,amsthm,mathtools,bm,bbm}
\usepackage{microtype}
\usepackage{enumitem}
\usepackage{algorithm}
\usepackage{algpseudocode}
\usepackage{booktabs}
\usepackage{hyperref}
\hypersetup{colorlinks=true,linkcolor=blue,citecolor=blue,urlcolor=blue}
\usepackage[numbers]{natbib}
\usepackage{tikz}
\usepackage{xcolor}
\usepackage{comment}
\usepackage{algpseudocode}

\usetikzlibrary{arrows.meta,positioning,fit}
\usepackage{xcolor}
\usetikzlibrary{positioning,fit}

\newcommand{\jun}[1]{\textcolor{red}{[Jun: #1]}}
\definecolor{TinaCrimson}{HTML}{DC143C} 
\newcommand{\titlefont}{\centering\color{TinaCrimson}\normalfont\bfseries\fontsize{16}{19}\selectfont}

\theoremstyle{plain}
\newtheorem{theorem}{Theorem}

\newtheorem{lemma}[theorem]{Lemma}
\newtheorem{corollary}[theorem]{Corollary}
\theoremstyle{definition}
\newtheorem{definition}[theorem]{Definition}
\newtheorem{assumption}[theorem]{Assumption}
\theoremstyle{remark}
\newtheorem{remark}[theorem]{Remark}

\DeclareMathOperator*{\argmax}{argmax}

\title{\titlefont{Memento 2: Learning by Stateful Reflective Memory}}
\author{
Jun Wang\\
\textit{UCL Centre for Artificial Intelligence}\\
\texttt{jun.wang@ucl.ac.uk}
}
\date{}

\begin{document}
\maketitle

\begin{abstract}
We present a theoretical study of continual and experiential learning in Large Language Model (LLM) agents that integrate episodic memory with reinforcement learning. We identify \emph{reflection}, the ability of an LLM to leverage past experience to inform future actions, as the key mechanism enabling continual adaptation without fine-tuning model parameters. Empirical results on Memento~\cite{zhou2025memento} and Case-Based Reasoning LLMs~\cite{guo2025optimizing,guo2024ds} show that episodic, experience-driven reflection supports generalised adaptation in a wide range of open-ended, long-horizon tasks, implying that efficient learning can occur during deployment and blurring the traditional separation between training and testing in machine learning.
Motivated by this observation, we introduce the
Stateful Reflective Decision Process (SRDP), a formal abstraction of reflective memory dynamics, where an agent maintains episodic memory and performs two key operations: \emph{writing}, which stores interaction outcomes and corresponds to \emph{policy evaluation}, and \emph{reading}, which retrieves relevant cases to make informed reflective decisions and corresponds to \emph{policy improvement}. This frames reflective memory as a control-theoretic object amenable to classical reinforcement learning analysis. We instantiate this \emph{Read--Write Reflective Learning} by integrating retrieval into soft policy iteration, with guaranteed convergence. We formally show that as memory expands and covers the state space more densely, the composite policy converges to the optimal solution. This theoretical framework unifies heuristic approaches such as case-based reasoning and retrieval-augmented generation with principled reinforcement learning and provides a rigorous mathematical basis for building reflective, memory-embedded LLM agents capable of continual general-purpose learning.
\end{abstract}

\begin{figure}[bh]
    \centering
    \includegraphics[width=1\textwidth]{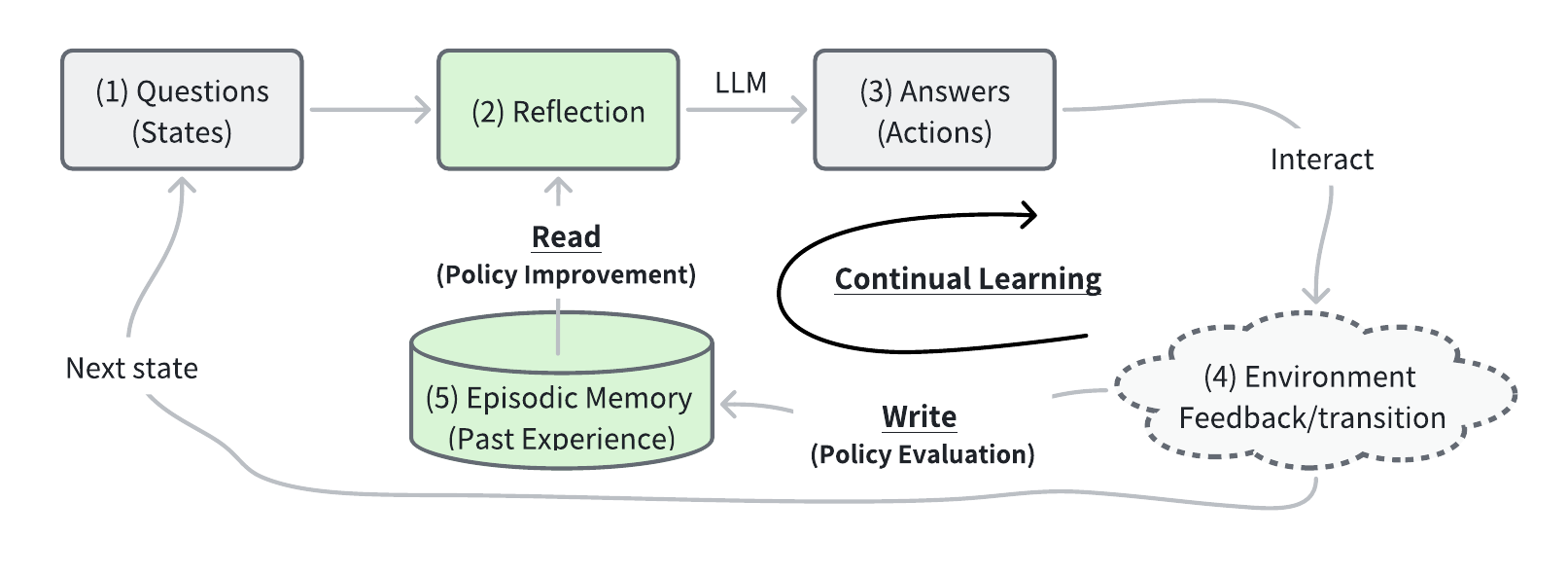}\vspace{-20pt}
    \caption{\emph{Learning by Stateful Reflection with Memory} realises {continual learning without fine-tuning LLMs} by iteratively 
\emph{reading} from (policy improvement) and \emph{writing} to (policy evaluation) an evolving 
\emph{episodic memory} that drives the agent’s online adaptation. Empirically, this \emph{Read-Write Learning} paradigm has shown strong effectiveness across diverse domains, including 
\emph{software testing}~\cite{guo2025optimizing}, \emph{automated data science}~\cite{guo2024ds}, 
and \emph{deep research agents}~\cite{zhou2025memento}, demonstrating that memory-driven reflection 
can endow LLM agents with genuine self-improving capabilities.
}
    \label{fig:cbr}
\end{figure}

\newpage

\section{Introduction}


Modern machine learning is about learning from experience \cite{turing1996intelligent,silver2025welcome}; both supervised learning and reinforcement learning can be viewed as processes of optimising a parameterised model through experience, either from human-labelled examples or data-driven feedback. The objective is to learn a mapping from inputs to outputs, or in the case of reinforcement learning, from states to actions, such that the resulting function generalises effectively to new, unseen situations \cite{shalev2014understanding,valiant1984theory}. Formally, the learner assumes that the target function is constrained by a family of parameterised models, for instance, neural networks, where the parameters determine the precise form of the mapping. 

In supervised learning, these parameters are typically adjusted by minimising a loss function defined over labelled examples, whereas in reinforcement learning (RL), the supervision signal is indirect: instead of labelled outputs, the agent interacts with an environment, receives rewards, and adjusts its parameters to maximise the expected cumulative return \cite{SuttonBarto2018}.
In both cases, the parameters are updated by backpropagation, a gradient-based optimisation algorithm that propagates errors through the network to refine the model \cite{rumelhart1986learning}.



Despite their substantial empirical successes, current machine learning paradigms suffer from fundamental inefficiencies in sample complexity. Learning via backpropagation and gradient-based optimisation typically requires an enormous number of training examples, whether labelled data in supervised learning or interaction trajectories in reinforcement learning, before competent behaviour emerges \cite{goodfellow2016deep,guo2025deepseek}. From a theoretical perspective, this reflects the difficulty of identifying generalisable structure in high-dimensional, non-linear hypothesis classes when learning is driven almost entirely by statistical correlations \cite{shalev2014understanding}. In the absence of explicit access to semantic abstractions, compositional structure, or task-level meaning, learning algorithms must recover regularities indirectly through averaging over large numbers of samples, leading to intrinsically high sample complexity.

The training of large language models continuously exemplifies this frustration. Although massive pre-training endows them with broad linguistic and reasoning abilities, adapting them to specific tasks or domains still has to resort to additional fine-tuning, such as supervised fine-tuning (SFT) or reinforcement learning with human feedback (RLHF) \cite{ouyang2022training}, each demanding further large quantities of curated data or human annotations \cite{lightman2023let}. 

In contrast, humans leverage semantic representations, episodic memory, and reflective reuse of experience to adapt with far fewer examples \cite{GershmanDaw2017EpisodicRL}. This contrast suggests that learning mechanisms beyond purely statistical optimisation, based on stateful memory and semantic reflection, may offer a principled path to lower complexity learning without large-scale gradient updates.

\begin{figure}[t]
    \centering
    \includegraphics[width=1\textwidth]{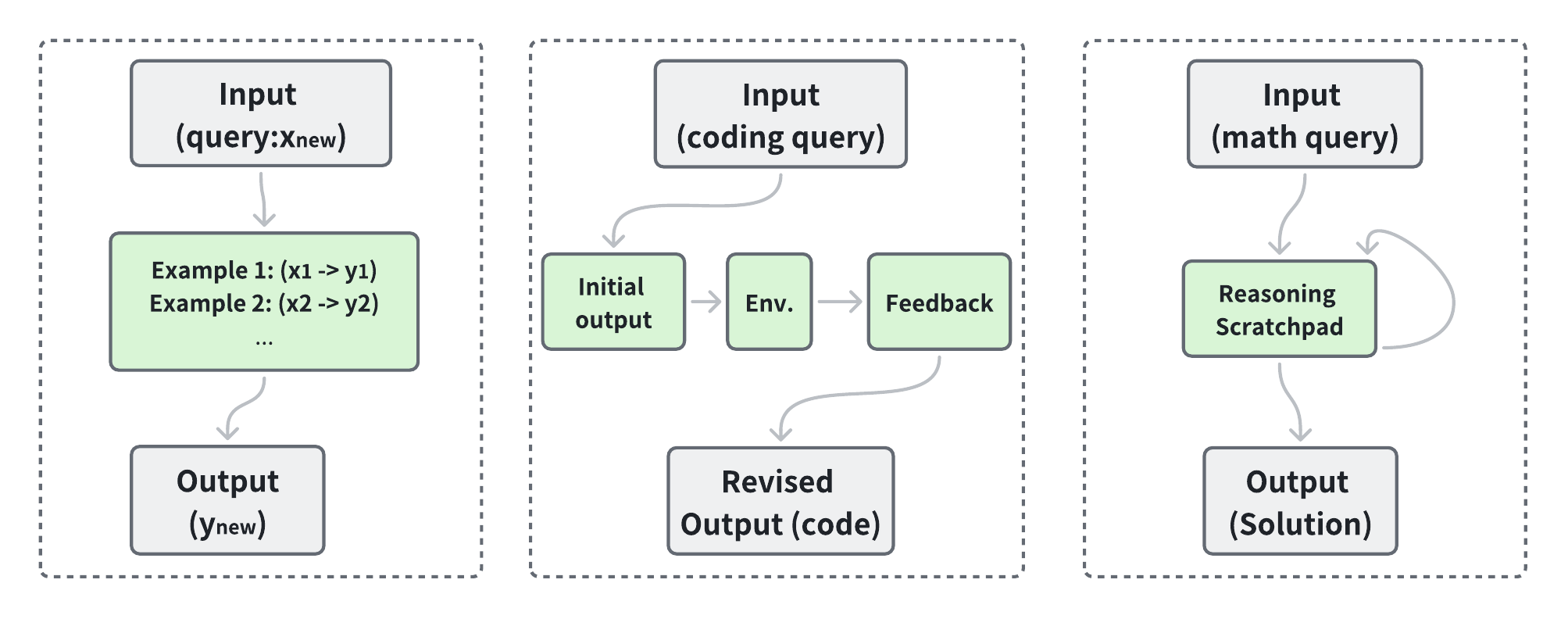}\vspace{-10pt}
    (a) In-context learning  \ \ \       (b) Feedback-driven reflection   \ \ \ \       (c) Chain-of-thought
    \caption{Three modes of \emph{generalised} reflection in LLM agents, where reflection broadly refers to information gathering or internal computation conducted before an answer is produced: (a) in-context learning where few-shot exemplars in the prompt steer behaviour on new tasks; (b) post-output reflection via environmental feedback; (c) internal chain-of-thought and planning before emitting an answer.}
    \label{fig:three-reflections}
\end{figure}

Interestingly, despite remaining data hungry and inefficient to train, current LLMs exhibit early signs of human like cognitive abilities that support iterative generalisation to unseen tasks from only a small number of examples. These emerging abilities can be broadly categorised into three forms of what we term \emph{generalised reflection}, as illustrated in Figure~\ref{fig:three-reflections}:

\begin{enumerate}[leftmargin=*]
    \item \textbf{In-context learning.}  
    As early as GPT-3, researchers observed that LLMs could learn from a few human-provided examples directly within a prompt, a process known as \emph{in-context learning}, where a few labelled examples guide behaviour on new tasks~\cite{brown2020language} (Figure~\ref{fig:three-reflections} (a)).  For instance, given a few examples of English-to-French translations in the prompt, GPT-3 can correctly translate a new English sentence into French without any parameter updates.

    \item \textbf{Feedback-driven reflection.}  
    Subsequent studies showed that given a specific unknown task, LLMs could refine their outputs through \emph{interaction with the environment}, adjusting responses based on external feedback~\cite{Yao2022ReAct} (Figure~\ref{fig:three-reflections} (b)).  For instance, in code generation, the LLM proposes a function, runs tests to observe errors or runtime feedback, and iteratively refines the code to reach a correct solution.

    \item \textbf{Internal reasoning.}  
    More recently, LLM models have displayed the ability to perform \emph{internal reasoning} via \emph{chain-of-thought} processes before generating final outputs~\cite{Wei2022CoT} (Figure~\ref{fig:three-reflections} (c)).  For instance, when solving a math word problem, the model explicitly writes intermediate reasoning steps (e.g., decomposing a problem into smaller parts) before computing the final numerical answer.
\end{enumerate}

Although these mechanisms vary in complexity from human-guided examples (Figure~\ref{fig:three-reflections} (a)) to environment-driven feedback (Figure~\ref{fig:three-reflections} (b)) to purely internal reasoning (Figure~\ref{fig:three-reflections} (c)), they share a common principle: each relies on \emph{intermediate reflection} such as information gathering or internal computation to enable adaptation and generalisation beyond the training data. 
Such generalised reflection has been observed not only in pre-trained large language models (LLMs)~\cite{brown2020language,guo2025deepseek}, but also in multimodal LLMs~\cite{driess2023palm} and non-language transformers~\cite{hollmann2025accurate,yu2025self}.

In our earlier empirical work \cite{guo2024ds,guo2025optimizing,zhou2025memento}, we investigated whether reflection through memory could enable learning without parameter updates. We first introduced this paradigm with \emph{case-based reasoning} LLM agents~\cite{guo2024ds}, where adaptation arises from updating memory rather than fine-tuning model weights. Building on this foundation, \emph{Memento}~\cite{zhou2025memento} formalised the idea within a \emph{Memory-Augmented Markov Decision Process (M-MDP)} framework. Across domains such as \emph{software testing}~\cite{guo2025optimizing}, \emph{automated data science}~\cite{guo2024ds,Grosnit2024KolbBasedEL}, and \emph{deep research}~\cite{zhou2025memento}, this memory-driven reflection paradigm has consistently shown that LLMs can achieve continual self-improvement through accumulated experience rather than model retraining.

In this paper, we further provide a theoretical analysis and formalisation and leverage the capability of \emph{generalised reflection} to develop a new paradigm for sample-efficient learning from experience without relying on data-intensive model training. 
Instead of updating model parameters, the proposed learning process operates through \emph{intermediate reflection}, driven by relevant past experiences retrieved from an episodic memory~\cite{GershmanDaw2017EpisodicRL}, as shown in Fig~\ref{fig:cbr}.
This framework can unify diverse sources of experience, whether originating from internal reasoning (e.g., chain-of-thought), interactions with the environment (e.g., ReAct~\cite{Yao2022ReAct}), or human-provided labelled data, under a single, memory-based learning mechanism.

We establish a theory of redefining how learning and decision-making are modelled through a proposed Stateful Reflective Decision Process
(SRDP). Unlike the traditional Markov Decision Process (MDP), our framework introduces several conceptual shifts. First, the agent’s state now combines the current situation with its episodic memory, allowing it to incorporate past experiences directly while still maintaining a coherent Markov structure in this enlarged space (namely \emph{reflected MDP}). Second, each action unfolds in two stages: the agent first recalls a relevant past case from memory, and then the language model produces a concrete response or behaviour based on that case. Third, the environment’s transitions and rewards are defined through the probabilistic reasoning of the language model itself, meaning that the model’s internal uncertainty is embedded in how the world evolves. Finally, memory evolves over time as the agent writes new experiences into it. This continual update makes the system dynamic and self-adaptive, without violating the overall Markovian formulation.

Within this formulation, learning naturally emerges from the process of retrieval: policy improvement is achieved through more effective retrieval, while policy evaluation is realised through writing experiences back to memory. 
This mechanism closely mirrors the principles of episodic control and reinforcement learning in humans~\cite{GershmanDaw2017EpisodicRL,fountas2024human}. 
Consequently, such learning can be efficiently implemented using retrieval-augmented generation architectures equipped with large-scale memory systems~\cite{Borgeaud2022RETRO,Lewis2020RAG}.

A key desired property is its self-learnability. With a frozen LLM\footnote{While LLM parameters may evolve over time through offline updates given adequate data, such updates typically occur at a much slower timescale. Our approach does not rule out parameter learning, but instead addresses the intermediate regime in which adaptation must occur without immediate model updates.} as its reasoning and reflection core, the agent improves through trial and error when data is limited and later reuses and combines past knowledge as experience grows. It balances exploitation (using known cases), exploration (targeted search in uncertain areas), and discovery (seeking new ones) through adaptive memory operations, achieving continual learning without parameter updates.

The paper is organised as follows. We first introduce the stateful reflective decision process and its Read-Write Reflective Learning. We then present theoretical results on convergence and analyse the model’s optimality and its properties as a continuously-learnable agent. Related work is reviewed next, and the paper concludes with directions for future research.

\section{Stateful Reflection as Generalised Learning Methods}

Recent advances in reflective prompting, such as Tree of Thoughts \cite{yao2023tree} and Graph of Thoughts \cite{besta2024graph}, have demonstrated the potential of structured reflection to enhance LLM reasoning. From a different perspective, we explore a new learning paradigm that enables models to learn from experience through reflection grounded in \emph{episodic memory}, rather than through parameter fine-tuning \cite{guo2024ds,guo2025optimizing,zhou2025memento}. To formalise this paradigm, we introduce the \emph{Stateful Reflective Decision Process} (SRDP), which models continual learning as a sequence of reflective interactions with memory. 

We next present the core concepts, with the full mathematical formulation deferred to later sections. Covering the three modes of generalised reflection introduced earlier (Fig.~\ref{fig:three-reflections}), we define \emph{reflection} in this work as follows.

\begin{definition}[Reflection]
Reflection (generalised) is an iterative mechanism by which a large language model (LLM) agent progressively improves its effective policy through coordinated interaction with episodic memory, internal reasoning processes, and environmental feedback. It is not a single operation but a two-step procedure consisting of a \emph{read} step followed by a \emph{write} step, which together form a closed learning loop.
\end{definition}



Figure~\ref{fig:cbr} illustrates the overall workflow of an SRDP. We begin by formally defining
\emph{episodic memory} as a structured repository of past interaction episodes, consistent
with the classical notion of episodic memory in cognitive psychology~\cite{tulving1983elements}.

\begin{definition}[Episodic Memory]
An episodic memory $M_t$ is a finite (and growing) collection of memory items
\[
M_t = \{ m_i \}_{i=1}^{N_t}, \qquad
m_i := (s_i, a_i, r_i, s'_i),
\]
where each item $m_i$ records a past interaction consisting of the environment
state $s_i \in S$, the executed action $a_i \in A$, the received reward
$r_i \in \mathbb{R}$, and the subsequent state $s'_i \in S$.
The space of all finite episodic memories is denoted by $\mathfrak{M}$.
Episodic memory evolves over time through a write operation
\[
{M}_{t+1} = \mathrm{Write}({M}_t, s_t, a_t, r_t, s_{t+1}).
\]
\end{definition}


Given a task or query represented as a state $s$~(1), the agent enters an internal reflection stage~(2), during which the LLM performs a read operation on ${M}$ to retrieve relevant episodic memory items or their summaries. Conditioned on this retrieved information, the LLM selects an external action or generates an answer $a$~(3), which is executed in the environment~(4) and produces evaluative feedback, including rewards, correctness signals, or task level assessments. Finally, the resulting feedback is incorporated into episodic memory via a write operation~(5), which stores new interaction episodes. 


Formally, let $\pi_t$ denote the agent’s effective policy at iteration $k$, implicitly represented by its prompt, episodic memory contents, and internal reasoning configuration. Reflection defines an operator $\mathcal{T}$ such that
\[
\pi_{t+1} = \mathcal{T}(\pi_t) = \text{\emph{Read}}\bigl(\text{\emph{Write}}(\pi_t)\bigr).
\]
The \emph{Read} step corresponds to policy improvement: the agent conditions on retrieved episodic memories, contextual exemplars, or intermediate reasoning traces to construct an improved action distribution for the current state,
\[
\pi_{t} \leftarrow \text{\emph{Read}}(\pi_{t-1} \mid s_t, M_t),
\]
where $s_t$ denotes the current environment state and $M_t$ the episodic memory state. In LLM agents, such conditioning reshapes the effective policy directly, without parameter updates.

The \emph{Write} step corresponds to policy evaluation: the agent generates an action, plan, or answer under $\pi_{t+1}$, producing trajectories
\[
\tau_t := m_i = (s_i, a_i, r_i, s'_i) \sim \pi_{t},
\]
whose outcomes yield evaluative feedback,
\[
\hat{V}(\pi_{t+1}) \leftarrow \text{Eval}(\tau_t),
\]
which is subsequently stored in episodic memory for future reads,
\[
M_{t+1} \leftarrow \text{\emph{Write}}(M_t,\tau_t \sim \pi_{t}).
\]

As we shall show in the following sections, this alternating reflective read--write reflective mechanism is formally analogous to \emph{policy iteration} in reinforcement learning. Policy improvement is realised through memory-conditioned inference, while policy evaluation is realised through interaction-driven feedback and experience accumulation. Under mild assumptions that reading yields non-decreasing policy quality with respect to accumulated feedback and that writing provides informative evaluation signals, the iterative reflection process converges to a fixed point corresponding to an optimal or self-consistent policy.

SRDP generalises and unifies several existing learning paradigms.
Compared with case based reasoning (CBR), which retrieves and adapts discrete past cases for problem solving \cite{guo2024ds,AamodtPlaza1994}, SRDP embeds retrieval within a sequential decision making framework. This explicitly models temporal dependencies, memory evolution, and long horizon effects under a Markovian assumption, rather than treating cases as isolated problem instances.

In contrast to retrieval augmented generation (RAG), which conditions generation on static external documents or databases \cite{Borgeaud2022RETRO}, SRDP treats memory as a dynamic and stateful component. Episodic memory is continually updated through interaction with the environment and grounded in evaluative feedback, allowing retrieval to reflect the agent’s own experience rather than fixed external knowledge. From the perspective of memory based reinforcement learning, SRDP replaces parameter level updates with reflective memory updates mediated by an LLM. Policy improvement is achieved through structured recall and reflection over episodic memory rather than gradient based optimisation of model parameters \cite{ramani2019short}. This shift enables continual adaptation while keeping the underlying LLM fixed, positioning reflection as the primary learning mechanism.

This formulation is also strongly inspired by neuroscientific findings on the role of the hippocampus in episodic memory and reflective cognition \cite{hami2025,McClelland1995CLS}. In humans and animals, episodic recall supports flexible reasoning and planning by replaying past experiences, an ability mirrored by the reflective cycles of SRDP. Thus, the Stateful Reflective Decision Process provides both a computational and cognitive bridge: it formalises how memory and reflection together enable continual, self-improving intelligence without explicit retraining.

Experimentally, it has been demonstrated that this approach enables learning without fine-tuning across various agentic tasks, including data-science agents~\cite{guo2025optimizing}, software-testing agents~\cite{guo2024ds}, and deep-research agents~\cite{zhou2025memento}.

\section{A Model for Reflective Process over Episodic Experience}
Mathematically, the core idea is to extend the Markov Decision Process \cite{Puterman1994} with an intermediate reflection stage, where the agent retrieves relevant experiences from external memory before generating the final output, enabling continual and memory-based learning. 

\subsection{Preliminary: Markov Decision Process}
\label{sec:mdp}
To make self-contained, we begin by reviewing the Markov Decision Process (MDP) \cite{Puterman1994}, the standard framework for modelling decision-making in agents.
An LLM agent, viewed as a large language model capable of reasoning and acting through interaction with its environment \cite{christianos2023pangu,yang2025agentic,wang2025tutorial,wan2024alphazero}, can be naturally described within this framework.
Formally, an MDP is defined by the tuple \[\mathcal{D} = \langle \mathcal{S}, \mathcal{A}, \mathcal{P}, \mathcal{R}, \gamma \rangle,\]where $\mathcal{S}$ is the state space, $\mathcal{A}$ is the action space, and 
$\mathcal{P}(s' \mid s,a)$ denotes the transition kernel, specifying the probability of reaching state $s'$ after taking action $a$ in state $s$;
$\mathcal{R}(s,a) \in \mathbb{R}$ represents the expected immediate reward;
and $\gamma \in [0,1)$ is the discount factor. Here, the \emph{state} $s$ corresponds to the current prompt or task context. Specifically, in the context of LLM reasoning, states correspond to intermediate reasoning steps \cite{wang2025tutorial,wang2024openr}, and when interacting with the external environment, they may take the form of textual descriptions of the environment state \cite{christianos2023pangu}.
The \emph{action} $a$ is the model’s generated output or decision,
and the \emph{transition} describes how the environment or subsequent prompt evolves after the model’s response \cite{wang2025tutorial}.
A (stochastic) policy $\pi(a \mid s)$ defines a probability distribution over actions conditioned on the current state and, in turn, induces the value function that measures the expected return when following the policy from a given state
\[
V^{\pi}_{\mathcal{D}}(s) =
\mathbb{E}_{\pi,\mathcal{P}}\!\left[
\sum_{t=0}^{\infty} \gamma^t \mathcal{R}(s_t,a_t)
\,\bigg|\, s_0 = s
\right],
\]
with the state–action value function
\[
Q^{\pi}_{\mathcal{D}}(s,a) = \mathcal{R}(s,a)
+ \gamma\,\mathbb{E}_{s'\sim\mathcal{P}(\cdot\mid s,a)}\!\left[V^{\pi}_{\mathcal{D}}(s')\right].
\]

The optimal value functions are
\[
V^\star_{\mathcal{D}}(s) = \max_{\pi} V^{\pi}_{\mathcal{D}}(s),
\qquad
Q^\star_{\mathcal{D}}(s,a) = \mathcal{R}(s,a)
+ \gamma\,\mathbb{E}_{s'\sim\mathcal{P}(\cdot\mid s,a)}\!\left[\max_{a'} Q^\star_{\mathcal{D}}(s',a')\right].
\]

For large language model (LLM) agents \cite{christianos2023pangu,wan2024alphazero,zeng2024token}, the policy is typically given by the LLM itself, which maps input states or prompts directly to action distributions through its generative reasoning process:
\[
\pi_{\mathrm{LLM}}(a \mid s)  \triangleq  p_{\mathrm{LLM}}(a \mid s;\,\theta),
\]
where $\theta$ denotes the (frozen in our case) parameters of the pre-trained model. 
Next, we extend the LLM’s stochastic policy by incorporating contributions from episodic memory, allowing past experiences to influence current decisions.

\subsection{Stateful Reflective Decision Process}

Building on the standard Markov Decision Process (MDP) framework introduced in the previous section, we now present the proposed \emph{Stateful Reflective Decision Process (SRDP)}. The classical MDP $\langle \mathcal{S}, \mathcal{A}, \mathcal{P}, \mathcal{R}, \gamma \rangle$ captures agent--environment interaction in its simplest form but lacks mechanisms for reflection and memory. To enable reflective, memory-driven behaviour, we extend this formulation by augmenting the agent with an episodic memory space $ M$, a memory retrieval policy $\mu$, and a stochastic LLM kernel $p_{\mathrm{LLM}}$ that generates context-dependent actions conditioned on both the current environment state and retrieved memory, yielding the Stateful Reflective Decision Process (SRDP). 
\subsubsection{Definition}

\begin{definition}[Stateful Reflective Decision Process (SRDP)]
\label{def:llm-mmdp}
SRDP is a tuple
\[
\mathcal{D}_{\mathrm{SRDP}}=\langle \mathcal{S}, \mathcal{A}, \mathcal{P}, \mathcal{R}, \gamma, \mathfrak{M}, p_{\mathrm{LLM}} \rangle
\]
that extends the standard MDP formalism by incorporating episodic memory and a language-model-based decision kernel. The components are defined as follows:
\begin{itemize}[leftmargin=*]
    \item $\mathcal{S}$: state space; 
    \item $\mathcal{A}$: environment action space;
    \item $\mathcal{P}(\cdot\mid s,a)$: transition kernel;
    \item $\mathcal{R}(s,a)\in\mathbb{R}$: reward function;
    \item $\gamma\in[0,1)$: discount factor;
    \item $\mathfrak{M}$: space of finite episodic memories, where each memory $M\in\mathfrak{M}$ is a multiset of cases $c=(s,a,r,s')$\footnote{In multi-state settings, $r$ can be replaced by an estimated value $Q$, where Bellman or TD updates are propagated across memory.};
    \item $p_{\mathrm{LLM}}(a\mid s,c)$: the LLM kernel that generates an environment action conditioned on the current state $s$ and a retrieved case $c$.
\end{itemize}

At time $t$, the agent maintains episodic memory $M_t=\{m_i\}_{i=1}^{N_t}$, where $N_t$ is the number of cases in the current memory, and observes the current environment state $s_t$. The decision process unfolds in two stages:
\begin{enumerate}[leftmargin=*]
    \item \emph{Retrieval stage:} A retrieval action $c_t$ is sampled from the current
episodic memory according to the retrieval policy
\[
c_t \sim \mu(\cdot \mid s_t, M_t),
\]
where $\mu$ may be similarity-based or learned. The action set is memory-dependent (indexer):
for each $M \in \mathfrak M$, $\mathcal C(M) := M$, and each $c \in \mathcal C(M)$
indexes a memory case used for retrieval.

    \item \emph{Action Stage:} The LLM generates an action conditioned on the retrieved case:
    \[
    a_t \sim p_{\mathrm{LLM}}(\cdot \mid s_t, c_t).
    \]
\end{enumerate}

This defines a \emph{composite policy}:
\begin{equation}
\pi^{\mu}(a_t \mid s_t, M_t) = \sum_{c_t \in M_t} \mu(c_t \mid s_t, M_t)\, p_{\mathrm{LLM}}(a_t \mid s_t, c_t).
\label{eq:composite-policy}
\end{equation}

After executing the action $a_t$, the environment changes to $s_{t+1}\sim\mathcal{P}(\cdot\mid s_t,a_t)$, yields a reward $r_t :=\mathcal{R}(s_t,a_t)$, and the memory updates to $M_{t+1}=\mathsf{Write}(M_t,s_t,a_t,r_t,s'_t)$.

\end{definition}




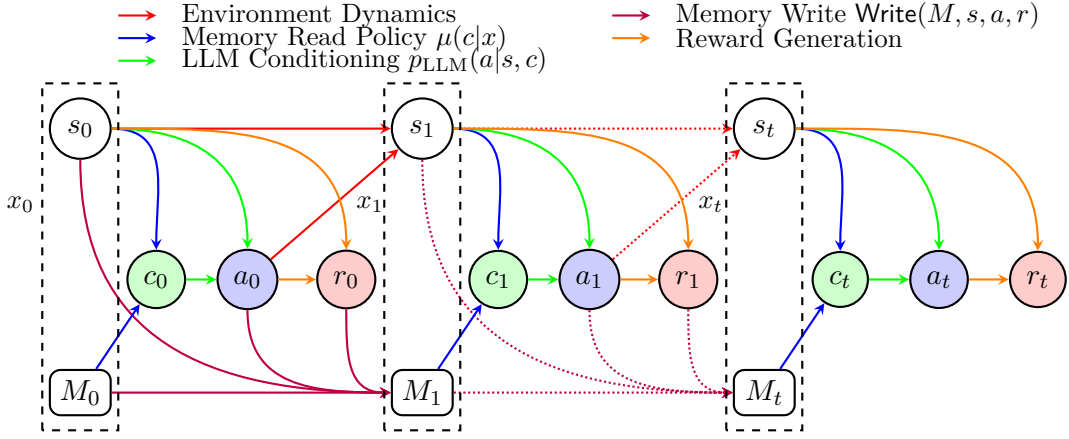
\begin{figure}[t]
\centering
\begin{tikzpicture}[
  node distance=1.5cm,
  state/.style={circle, draw, thick, minimum size=0.8cm},
  memory/.style={rectangle, rounded corners, draw, thick, minimum width=0.8cm, minimum height=0.6cm},
  action/.style={circle, draw, thick, minimum size=0.7cm, fill=blue!20},
  reward/.style={circle, draw, thick, minimum size=0.7cm, fill=red!20},
  retrieval/.style={circle, draw, thick, minimum size=0.7cm, fill=green!20},
  >=stealth
]

\node[state] (s0) at (-1,4) {$s_0$};
\node[memory] (M0) at (-1,0.5) {$M_0$};
\node[retrieval] (c0) at (0,2) {$c_0$};
\node[action] (a0) at (1.2,2) {$a_0$};
\node[reward] (r0) at (2.5,2) {$r_0$};

\node[state] (s1) at (3.5,4) {$s_1$};
\node[memory] (M1) at (3.5,0.5) {$M_1$};
\node[retrieval] (c1) at (4.5,2) {$c_1$};
\node[action] (a1) at (5.7,2) {$a_1$};
\node[reward] (r1) at (7,2) {$r_1$};

\node[state] (st) at (8,4) {$s_t$};
\node[memory] (Mt) at (8,0.5) {$M_t$};
\node[retrieval] (ct) at (9,2) {$c_t$};
\node[action] (at) at (10.3,2) {$a_t$};
\node[reward] (rt) at (11.6,2) {$r_t$};


\draw[->, thick, red] (s0) -- (s1);
\draw[->, thick, red,densely dotted] (s1) -- (st);
\draw[->, thick, red] (a0) -- (s1);
\draw[->, thick, red,densely dotted] (a1) -- (st);

\draw[->, thick, purple] (M0) -- (M1);
\draw[->, thick, purple,densely dotted] (M1) -- (Mt);

\draw[->, thick, purple] (s0) to[out=270, in=180, looseness=1.2] (M1);
\draw[->, thick, purple] (a0) to[out=270, in=180, looseness=1.2] (M1);
\draw[->, thick, purple] (r0) to[out=270, in=180, looseness=1.2] (M1);

\draw[->, thick, purple,densely dotted] (s1) to[out=270, in=180, looseness=1.2] (Mt);
\draw[->, thick, purple,densely dotted] (a1) to[out=270, in=180, looseness=1.2] (Mt);
\draw[->, thick, purple,densely dotted] (r1) to[out=270, in=180, looseness=1.2] (Mt);

\draw[->, thick, blue] (s0) to[out=0, in=90, looseness=1.2] (c0);
\draw[->, thick, blue] (M0) -- (c0);
\draw[->, thick, blue] (s1) to[out=0, in=90, looseness=1.2] (c1);
\draw[->, thick, blue] (M1) -- (c1);
\draw[->, thick, blue] (st) to[out=0, in=90, looseness=1.2] (ct);
\draw[->, thick, blue] (Mt) -- (ct);

\draw[->, thick, green] (s0) to[out=0, in=90, looseness=1.2] (a0);
\draw[->, thick, green] (c0) -- (a0);
\draw[->, thick, green] (s1) to[out=0, in=90, looseness=1.2] (a1);
\draw[->, thick, green] (c1) -- (a1);
\draw[->, thick, green] (st) to[out=0, in=90, looseness=1.2] (at);
\draw[->, thick, green] (ct) -- (at);

\draw[->, thick, orange] (s0) to[out=0, in=90, looseness=1.2] (r0);
\draw[->, thick, orange] (a0) -- (r0);
\draw[->, thick, orange] (s1) to[out=0, in=90, looseness=1.2] (r1);
\draw[->, thick, orange] (a1) -- (r1);
\draw[->, thick, orange] (st) to[out=0, in=90, looseness=1.2] (rt);
\draw[->, thick, orange] (at) -- (rt);

\draw[thick, dashed] (-1.5,0) rectangle (-0.5,4.6);
\node[anchor=west] at (-2.1,3) {\small $x_0$};

\draw[thick, dashed] (3,0) rectangle (4,4.6);
\node[anchor=west] at (2.5,3) {\small $x_1$};

\draw[thick, dashed] (7.5,0) rectangle (8.5,4.6);
\node[anchor=west] at (7,3) {\small $x_t$};

\draw[->, thick, red] (-0.5,5.5) -- (0,5.5);
\node[anchor=west] at (0.2,5.5) {\small Environment Dynamics};

\draw[->, thick, blue] (-0.5,5.2) -- (0,5.2);
\node[anchor=west] at (0.2,5.2) {\small Memory Read Policy $\mu(c|x)$};

\draw[->, thick, green] (-0.5,4.9) -- (0,4.9);
\node[anchor=west] at (0.2,4.9) {\small LLM Conditioning $p_{\mathrm{LLM}}(a|s,c)$};

\draw[->, thick, purple] (6,5.5) -- (6.5,5.5);
\node[anchor=west] at (6.7,5.5) {\small Memory Write $\mathsf{Write}(M,s,a,r)$};

\draw[->, thick, orange] (6,5.2) -- (6.5,5.2);
\node[anchor=west] at (6.7,5.2) {\small Reward Generation};

\end{tikzpicture}
\caption{A graphical model of the Stateful Reflective Decision Process (SRDP) showing the augmented state space $x_t = (s_t, M_t)$, retrieval actions $c_t$, LLM conditioning, and memory evolution. Dashed boxes indicate augmented states combining environment state and memory.}
\label{fig:m-mdp-graphical}
\end{figure}

Figure~\ref{fig:m-mdp-graphical} depicts the graphical model of the Stateful Reflective Decision Process (SRDP), characterised by a \emph{dual-action structure}.
The agent first performs a \emph{retrieval action} $c_t \in M_t$ (green nodes), selecting a relevant past case from memory to guide its current decision.
This retrieval is controlled by the policy $\mu(c \mid s, M)$, represented by the blue arrows connecting the current environment state $s_t$ and memory $M_t$ to the retrieval node $c_t$.
The policy evaluates both the similarity between the current state and stored experiences, and the local density of memory cases.
Conditioned on the retrieved case, the LLM then reflects upon the retrieved experience and generates the environment action $a_t \in \mathcal{A}$ (blue nodes) via its generative kernel $p_{\mathrm{LLM}}(a \mid s, c)$, shown by the green arrows.
Through this hierarchical process, the retrieval stage provides contextual grounding, while the LLM generates the final action, jointly realising \emph{stateful reflection}, as illustrated in Figure~\ref{fig:cbr}.

A key advantage of this dual-action structure is that it defines a composite policy (Eq.~\ref{eq:composite-policy}) that unifies classical reinforcement learning \cite{SuttonBarto2018} with LLM-based decision-making \cite{christianos2023pangu}, thereby combining the strengths of both paradigms.
To illustrate, consider two limiting cases.
In the first case, a naive LLM produces an action $a \in c$ directly from a retrieved case $c$, effectively treating the case as a prompt.
Here, the policy reduces to a standard reinforcement learning or episodic control scheme \cite{blundell2016model,pritzel2017neural,ramani2019short}, where the retrieval function $\mu$ selects the case $c$ with the highest $Q$-value, and the corresponding action is executed.
In the opposite extreme, when no relevant experience exists in memory, the agent must rely entirely on the internal knowledge of the LLM to generate actions.
Between these two cases lies the general setting of SRDP, where the agent can integrate both sources of information, the reasoning capability of the LLM and the experiential knowledge stored in memory, to produce more adaptive and effective decisions.

The memory evolution mechanism forms a crucial component of the SRDP dynamics. Purple arrows demonstrate how memory grows through $\mathsf{Write}(M,s,a,r)$ operations, with experience tuples $(s_t, a_t, r_t)$ being incorporated into the memory at the next time step. This creates a directed evolution $M_0 \to M_1 \to \cdots \to M_t$ where the memory size $|M_t|$ increases over time, expanding the agent's retrieval options. 

However, the introduction of memory fundamentally alters the dynamics of the system, making the process no longer strictly Markovian if the original state is considered. In a standard MDP, the next state depends only on the current state and action, i.e., the system satisfies
$
P(s_{t+1} \mid s_t, a_t, s_{t-1}, a_{t-1}, \ldots) = P(s_{t+1} \mid s_t, a_t).
$
However, once we introduce an evolving memory $M_t$ that accumulates past experience tuples $(s_i, a_i, r_i)$, the transition to the next augmented state depends on the entire history through $M_t$.
Formally,
$
P(s_{t+1}, M_{t+1} \mid s_t, a_t, M_t) \neq P(s_{t+1}, M_{t+1} \mid s_t, a_t)
$
because $M_t$ itself is a summary of all past interactions. We address this issue next by defining a simplified MDP that incorporates memory directly into the state representation while embedding the fixed LLM within the environment dynamics. This formulation enables us to focus on optimising the retrieval policy, with both the LLM’s behaviour and memory evolution captured within the new state–transition structure.

\subsubsection{Transforming to Reflected MDP}
To recover a Markovian formulation, we redefine the state to explicitly include memory,
forming an \emph{augmented state}, i.e., at time $t$, the agent occupies an augmented state
\[
x_t := (s_t,  M_t) \in \mathcal{X} := S \times \mathfrak{M}.
\]

The resulting process over $\{x_t\}$ is Markovian in this expanded space, even though the
environment process over $\{s_t\}$ alone is not. Each augmented state combines the current
environment state $s_t$ with the agent’s episodic memory $M_t$, as illustrated by the dashed
rectangles in Figure~\ref{fig:m-mdp-graphical}. Augmenting the state with memory restores the Markov property and enables a well-defined control formulation, i.e., for any reflection action $c_t$,
\[
\Pr(x_{t+1} \mid x_{0:t}, c_{0:t})
=
\Pr(x_{t+1} \mid x_t, c_t),
\]
where the environment transition depends only on $(s_t, a_t)$, while the memory update is a
deterministic or stochastic function of $(M_t, s_t, a_t, r_t,s'_t)$. Since the reflection mechanism
and the LLM action distribution are fixed functions of $(s_t, M_t)$, all sources of dependence
on the past are captured by $x_t$. Hence the augmented process satisfies the Markov property.

Since our focus is on learning through memory retrieval rather than modifying the LLM itself,
we treat the LLM as a fixed or slow-evolving component of the environment. Its action generation is absorbed
into the environment dynamics via the effective transition and reward functions
$\mathcal{P}^{\mathrm{LLM}}$ and $\mathcal{R}^{\mathrm{LLM}}$.
Under this view, the agent's only controllable decision is the retrieval action
\[
c_t \sim \mu(\cdot \mid s_t, M_t),
\]
while state transitions and rewards are determined jointly by the environment and the fixed
LLM kernel $p_{\mathrm{LLM}}(a \mid s_t, c_t)$.
The resulting Reflected MDP therefore treats the retrieval policy $\mu$ as the agent's primary
policy, with LLM mediated reasoning incorporated into the environment model. Specifically, we have:

\begin{definition}[Reflected MDP]
\label{def:llm-induced-mdp-v2}
The \emph{Reflected MDP} is a tuple
\[
\mathcal{D}_{\mathrm{ReMDP}}
= \langle \mathcal{X}, \mathcal{C}, \mathcal{P}_{\mathrm{LLM}}, \mathcal{R}_{\mathrm{LLM}}, \gamma \rangle,
\]
that transforms the underlying SRDP,  $\mathcal{D}_{\mathrm{SRDP}} = 
\langle \mathcal{S}, \mathcal{A}, \mathcal{P}, \mathcal{R}, \gamma, \mathfrak{M}, p_{\mathrm{LLM}} \rangle$, with the components are defined as follows:
\end{definition}
\begin{itemize}[leftmargin=*]
    \item \emph{Augmented state space:} $\mathcal{X} = \mathcal{S} \times \mathfrak{M}$, with each state $x = (s, M)$ consisting of the current environment state $s$ and the current episodic memory $M$.
    \item \emph{Action space:} $\mathcal C(M) :=M$, consisting of retrieval actions that index and select cases from the current episodic memory.
    \item \emph{Transition kernel:} For $x=(s,M)$ and $c\in\mathcal{C}$,
    \begin{equation}
    \mathcal{P}_{\mathrm{LLM}}(x' \mid x, c)
    = \sum_{a \in \mathcal{A}}
    p_{\mathrm{LLM}}(a \mid s, c)\,
    \mathbf{1}\{x' = (s', \mathsf{Write}(M, s, a, \mathcal{R}(s,a)))\}\,
    \mathcal{P}(s' \mid s, a),
    \label{eq:llm-transition}
    \end{equation}
    where $\mathsf{Write}(M, s, a, r)$ appends the new experience $(s,a,r)$ to memory.
    \item \emph{Reward function:}
    \begin{equation}
    \mathcal{R}_{\mathrm{LLM}}(x, c)
    = \sum_{a \in \mathcal{A}}
    p_{\mathrm{LLM}}(a \mid s, c)\,
    \mathcal{R}(s,a).
    \label{eq:llm-reward}
    \end{equation}
\end{itemize}

The \emph{Reflected MDP} provides a fully Markovian formulation over the augmented state space
$\mathcal{X} = \mathcal{S} \times \mathfrak{M}$, in which transitions and rewards correspond to
expected outcomes under the fixed LLM kernel $p_{\mathrm{LLM}}$. At each time step, the
transition unfolds as follows: the agent first retrieves a memory case
$c_t \sim \mu(\cdot \mid s_t, M_t)$; conditioned on this retrieval, the LLM generates an
environment action $a_t \sim p_{\mathrm{LLM}}(\cdot \mid s_t, c_t)$; the environment then
transitions to $s_{t+1} \sim \mathcal{P}(\cdot \mid s_t, a_t)$ and produces a reward
$\mathcal{R}(s_t, a_t)$; finally, episodic memory is updated according to
$M_{t+1} = \mathsf{Write}(M_t, s_t, a_t, r_t,s'_t)$.

By absorbing the retrieval and LLM generation mechanisms into effective transition and reward
functions over the augmented state space $\mathcal{X}$, this interaction induces a simplified
control process. A policy in the Reflected MDP is defined
as a retrieval policy \[\mu : \mathcal{X} \to \Delta(\mathcal{C}),\] which maps augmented states
$x=(s, {M})$ to probability distributions over retrieval actions. Environment actions are
generated implicitly by the fixed LLM kernel $p_{\mathrm{LLM}}(a \mid s, c)$ and absorbed into
the transition dynamics. This abstraction captures the full reflective behaviour of the agent in a compact Markovian form
that is amenable to classical reinforcement learning analysis and learning algorithms.

It is worth mentioning that this augmented state admits a natural cognitive interpretation. The environment state $s_t$
encodes the agent’s external sensory or task-level signal, while the episodic memory $M_t$
represents an internal cognitive state that stores and organises past experiences, outcomes,
and contextual information. The pair $(s_t, M_t)$ therefore constitutes the agent’s effective
decision state, integrating external observation with internal experience. This perspective
parallels belief-state constructions in partially observable Markov decision processes, where
internal memory summarises relevant history to support optimal control
\cite{kaelbling1998planning}, and aligns with cognitive theories that view episodic memory as a
core substrate for human decision making and learning \cite{tulving1983elements,GershmanDaw2017EpisodicRL}.

The underlying MDP captures the dynamics of the environment alone, independent of the agent’s internal mechanisms. In contrast, the SRDP explicitly models the coupled interaction among the agent, its episodic memory, and the LLM during decision making. The Reflected MDP then provides a Markovian reformulation of this process by lifting the SRDP to an augmented state space, in which the combined agent–memory–LLM dynamics become Markovian. The relationship between the underlying environment, the SRDP, and the resulting Reflected
MDP is summarised as

\begin{equation}
\underbrace{\langle \mathcal{S}, \mathcal{A}, \mathcal{P}, \mathcal{R}, \gamma \rangle}_{\text{Underlying MDP}}
\xrightarrow{\;\mathrm{ext.}\;}
\underbrace{\Big\langle \mathcal{S}, \mathcal{A}, \mathcal{P}, \mathcal{R}, \gamma,  M, \mu, p_{\mathrm{LLM}} \Big\rangle}_{\text{SRDP}}
\xrightarrow{\;\mathrm{equiv.}\;}
\underbrace{\Big\langle \mathcal{X} := \mathcal{S} \times  M, \mathcal{C}, \mathcal{P}^{\mathrm{LLM}}, \mathcal{R}^{\mathrm{LLM}}, \gamma \Big\rangle}_{\text{Reflected MDP}},
\label{three-mdps}
\end{equation}
where a detailed comparison between the SRDP and the Reflected MDP is provided in
Appendix~\ref{ap:comparison}.

\section{Read--Write Reflective Learning}
Under the reflected MDP abstraction, continual learning reduces to an iterative read--write reflective procedure over episodic memory, which we formalise as a learning algorithm next. 
\subsection{Policy Iteration}
We now define the retrieval mechanism that underpins the learning method.  
The central idea is to use \emph{local state density} to guide which past cases should be retrieved from memory.  
To capture the similarity structure in the memory bank, we employ \emph{Parzen window estimation} \cite{Parzen1962,wang2008unified}.

Let $K:\mathbb{R}^d\to\mathbb{R}_{\ge 0}$ be a smooth kernel (e.g., Gaussian) with bandwidth $h>0$, and let $\psi:\mathcal{S}\to\mathbb{R}^d$ be a state embedding. For memory $M=\{c_i=(s_i,a_i,r_i)\}_{i=1}^{N}$ and query state $s$, the Parzen similarity weight for case $c\in M$ is:
\begin{equation}
w_{\text{parzen}}(s,c) = \frac{K\left(\psi(s)-\psi(s(c))/h\right)}{\sum_{c'\in M}K\left(\psi(s)-\psi(s(c'))/h\right)},
\label{eq:parzen-weight}
\end{equation}
where $s(c)$ denotes the state component of a case $c$. 
The kernel defines a probability distribution over memory cases that places higher weight on states similar to the current query state $s$. Thus, the Parzen prior $ \mu_0(c\mid x) $ over retrieval actions at augmented state $x=(s,M)$ is:
\begin{equation}
\mu_0(c\mid x) \equiv w_{\text{parzen}}(s,c).
\label{eq:parzen-prior}
\end{equation}

In order to enable an LLM agent to act even when no relevant past cases exist in memory, 
we augment the case set $\mathcal{C}$ with a special \emph{void case}: $ M \leftarrow  M \cup \{c_{\varnothing}\}$. The void case $c_{\varnothing}$ represents the ability of the agent to generate an action directly from
its internal world knowledge, independently of retrieved memory. 
We formulate this by modifying the prior policy $\mu_0$ that incorporates a 
state-dependent mixture between memory-based retrieval and the void case. Let $K(x,c)$ be a similarity kernel for ordinary cases $c\in\mathcal{C}$, and let $K_{\varnothing}>0$ be a constant kernel score assigned to the void case $c_{\varnothing}$.  
Then, the prior distribution $\mu_0(\cdot\mid x)$ becomes
\begin{equation}
\label{eq:mu0-def}
\mu_0(c\mid x)
\equiv
\frac{
\begin{cases}
K(x,c), & c\neq c_{\varnothing}, \\[3pt]
K_{\varnothing}, & c = c_{\varnothing} ,
\end{cases}
}{
\sum_{c'\neq c_{\varnothing}} K(x,c') + K_{\varnothing} },
\end{equation}
where $K(x,c) := K\!\left(\frac{\psi(s(x))-\psi(s(c))}{h}\right)$.

It is useful to express this distribution as a mixture.  
Define the normalised memory-based distribution
\[
\mu_{\mathrm{mem}}(c\mid x)
=
\frac{K(x,c)}{\sum_{c'\neq c_{\varnothing}} K(x,c')},
\qquad c\neq c_{\varnothing},
\]
and let
\[
\lambda(x)
=
\frac{\sum_{c'\neq c_{\varnothing}}K(x,c')}{\sum_{c'\neq c_{\varnothing}}K(x,c') + K_{\varnothing}}.
\]

Then \eqref{eq:mu0-def} is equivalent to the mixture representation
\begin{equation}
\label{eq:mu0-mixture}
\mu_0(c\mid x)
=
\lambda(x)\,\mu_{\mathrm{mem}}(c\mid x)
+ \bigl(1-\lambda(x)\bigr)\,\delta_{c_{\varnothing}}(c),
\end{equation}
where $\delta_{c_{\varnothing}}$ denotes the unit point mass at $c_{\varnothing}$. This mixture form shows that the agent automatically interpolates between 
\emph{reflection(retrieval)-driven behaviour} when similarity to stored cases is high ($\lambda(x)\approx1$) and 
\emph{knowledge-driven discovery} through $c_{\varnothing}$ when similarity is low ($\lambda(x)\approx0$). 
This mechanism mirrors the ``discovery'' component in infinite-armed bandits~\cite{berry1997bandit}, 
complementing exploitation and exploration with the capacity to act beyond the current memory.
This mechanism ensures that the agent can always act, even in novel states, while still exploiting relevant past experiences when available.




We aim for the agent’s final policy $\mu(\cdot \mid x)$ to remain close to a prior policy $\mu_0$, while still being shaped by reward feedback from the environment. To formalise this trade-off, we define a KL-regularised objective in terms of the state–case value function $Q^\mu(x,c)$.
The KL-regularised Bellman \emph{evaluation} operator is
\begin{equation}
\label{eq:kl-eval-operator}
(\mathcal{T}^{\mu}_{\mathrm{KL}} Q)(x,c)
:=
\mathbb E_{a,s',M'}\!\left[
r
+
\gamma\left(
\sum_{c'} \mu(c'\mid x')\, Q(x',c')
-
\alpha\,\mathrm{KL}\!\bigl(
\mu(\cdot\mid x') \,\|\, \mu_{0}(\cdot\mid x')
\bigr)
\right)
\,\middle|\, x,c
\right],
\end{equation}
where $\mathcal{T}^\mu_{\mathrm{KL}} : \mathbb R^{\mathcal X \times \mathcal C}
\to \mathbb R^{\mathcal X \times \mathcal C}
$. It is worth emphasising that all value functions $Q(x,c)$ and $V(x)$ defined on the augmented
state space $\mathcal{X}$ of the Reflected MDP are distinct from the value functions
$Q^{\mathcal{D}}$ and $V^{\mathcal{D}}$ of the original MDP (see Section~\ref{sec:mdp}).
The latter are defined over the environment state space $S$, whereas the former explicitly
incorporate episodic memory and LLM-mediated action generation through the Reflected MDP
dynamics. We shall show that, by appropriately leveraging these LLM-induced dynamics,
the Reflected MDP can approximate the optimal value functions of the original MDP
in a more sample-efficient manner.

For any fixed $\mu$, this operator is a $\gamma$-contraction in the sup-norm and 
admits a unique fixed point $Q^\mu$, representing the KL-regularised value of~$\mu$.

\textbf{KL-Regularised Policy Improvement} 
Given a $Q$-function over retrieval actions, we define policy improvement as the solution to a KL-regularised optimisation problem.
Namely, at each augmented state $x$, the improved retrieval policy is:
\begin{equation}
\mu^{+}(c\mid x) \in \arg\max_{\nu\in\Delta(M)} \left\{ \sum_{c\in M}\nu(c|x)\,Q(x,c) - \alpha\,\mathrm{KL}(\nu(c|x)\|\mu_{0}(\cdot\mid x)) \right\},
\label{eq:parzen-kl-objective}
\end{equation}
where $\Delta(M)$ is the simplex over finite set, memory $M$, and $\alpha>0$ is the temperature parameter.

\begin{lemma}[also in \cite{neu2017unified}]\label{lemma-objective}
The optimisation problem \eqref{eq:parzen-kl-objective} has a unique solution:
\begin{equation}
\mu^{+}(c\mid x) = \frac{\mu_{0}(c\mid x)\,\exp(Q(x,c)/\alpha)}{\sum_{c'\in M}\mu_{0}(c'\mid x)\,\exp(Q(x,c')/\alpha)}.
\label{eq:parzen-kl-solution}
\end{equation}
\end{lemma}

The proof can be found in Appendix~\ref{appendix:proofs}. Because $\mu_0(c_{\varnothing}\mid x)>0$ for all $x$, the improved policy $\mu'$ always assigns 
nonzero probability to the void case, preserving the agent's capacity for discovery.

\textbf{KL-Regularised Policy Iteration}  Now, on the basis of policy evaluation in Eq.~(\ref{eq:kl-eval-operator}) and policy improvement in Eq.~(\ref{eq:parzen-kl-solution}), we are ready to give the policy iteration algorithm, as shown in Algorithm \ref{alg:kl-pi-fixed-memory}.
Under standard assumptions, this procedure yields a monotone improvement in the 
KL-regularised value and converges to the fixed point of the KL-regularised optimality operator, 
\[
(\mathcal{T}^*_{\mathrm{KL}}Q)(x,c)
=
\mathbb{E}\!\left[
r + \gamma\, \alpha\log\!\sum_{c'} 
\mu_0(c'\mid x')\,\exp\!\bigl(Q(x',c')/\alpha\bigr)
\,\middle|\, x,c
\right].
\]

\begin{algorithm}[t]
\caption{Soft Policy Iteration with Fixed Memory under Reflected MDP}
\label{alg:kl-pi-fixed-memory}
\begin{algorithmic}[1]
\State \textbf{Input:} Temperature $\alpha>0$, kernel bandwidth $h>0$
\State \textbf{Initialise:} Fixed episodic memory $M$, reference prior $\mu_0(\cdot\mid x)$
(including void case $c_{\varnothing}$), initial policy $\mu_0$, initial $Q_0$

\For{$t = 0,1,2,\ldots$}

\State \textbf{(Policy Evaluation)}
\State Compute the KL-regularised value function $Q^{\mu_t}$ as the unique fixed point of
\[
Q = \mathcal{T}^{\mu_t}_{\mathrm{KL}} Q,
\]
where (Eq.~(\ref{eq:kl-eval-operator}))
\[
(\mathcal{T}^{\mu_t}_{\mathrm{KL}} Q)(x,c)
=
\mathbb{E}\!\left[
r + \gamma
\Big(
\sum_{c'\in M} \mu_t(c' \mid x') Q(x',c')
-
\alpha \mathrm{KL}(\mu_t(\cdot \mid x') \| \mu_0(\cdot \mid x'))
\Big)
\,\middle|\, x,c
\right].
\]

\State \textbf{(Policy Improvement)}
\State Update the retrieval policy by solving
\[
\mu_{t+1}(\cdot \mid x)
\in
\arg\max_{\nu \in \Delta(M)}
\left\{
\sum_{c} \nu(c \mid x) Q^{\mu_t}(x,c)
-
\alpha \mathrm{KL}(\nu(\cdot \mid x)\|\mu_0(\cdot \mid x))
\right\},
\]
which admits the closed-form solution (Eq.~(\ref{eq:parzen-kl-solution}))
\[
\mu_{t+1}(c \mid x)
=
\frac{\mu_0(c \mid x)\exp(Q^{\mu_t}(x,c)/\alpha)}
{\sum_{c'} \mu_0(c' \mid x)\exp(Q^{\mu_t}(x,c')/\alpha)}.
\]

\EndFor
\end{algorithmic}
\end{algorithm}

For the convergence analysis, we start with assumptions followed with the theorem:
\begin{assumption}[Bounded rewards and discount]\label{as:bounded}
$|r|\le R_{\max}<\infty$, $\gamma\in[0,1)$.
\end{assumption}
\begin{assumption}[Stationary memory dynamics during evaluation]\label{as:stationary-memory}
Within each policy-evaluation phase, the memory kernel is stationary (equivalently, treat $M$ as part of the state).
\end{assumption}

\begin{theorem}[Convergence of Parzen-KL Soft Policy Iteration]
\label{thm:parzen-spi-convergence}
Under assumptions~\ref{as:bounded} and~\ref{as:stationary-memory} (bounded rewards and $\gamma < 1$ and stationary memory), the process (the reflected MDP) is an ordinary discounted MDP. Iterations between (\ref{eq:kl-eval-operator}) and (\ref{eq:parzen-kl-solution}) (Algorithm \ref{alg:kl-pi-fixed-memory}) converge to a fixed point $(Q^{\star},\mu^{\star})$ that is optimal for the KL-regularised objective with Parzen prior.
\end{theorem} 
We refer to Appendix for the proof. 

Some discussions: It is worth mentioning that the void case $c_{\varnothing}$ plays a crucial role: 
by providing a state-independent baseline probability in $\mu_0$, 
it ensures that policy iteration retains the ability to \emph{discover} new behaviours 
beyond exploitation or exploration of existing memory. Unlike classical density estimation, our memory cases are not independent and identically distributed but are gathered sequentially through policy execution. Moreover, the finite memory constraint means we perform kernel smoothing over a limited and evolving set of points rather than relying on asymptotic density estimation. As a result, bandwidth selection plays a critical role in balancing local similarity against statistical robustness as the memory continues to grow. Several strategies can be used to choose the bandwidth $h$. Silverman’s rule suggests setting $h \propto n^{-1/(d+4)}$, where $n=|M|$ is the memory size and $d$ the embedding dimension. Another approach is cross-validation, which tunes $h$ by optimising performance on held-out retrieval tasks. Finally, an adaptive strategy decreases $h$ as memory grows, ensuring that neighbourhoods remain local and informative.



\subsection{Two-Time-Scale Convergence with Online Memory Rewriting}
The preceding analysis assumes that episodic memory remains fixed. We now relax this
assumption and allow memory to evolve on a slower time scale. Let $\eta_t$ denote the step
size used for policy evaluation and improvement, and let $\rho_t$ denote the step size for
memory updates. We assume a two time scale regime in which
\[
\frac{\rho_t}{\eta_t} \to 0, \qquad
\sum_{t=1}^\infty \eta_t = \infty, \qquad
\sum_{t=1}^\infty \eta_t^2 < \infty.
\]

Under this separation of time scales, policy evaluation and improvement operate on a faster
time scale, while episodic memory evolves more slowly.  Formally, we make the following assumptions:



\begin{assumption}[Two-Time-Scale Conditions]
\label{as:twots-detailed}
\begin{enumerate}[label=(\roman*)]
\item \emph{Step sizes}: $\eta_t > 0$, $\rho_t > 0$ with:
\begin{align*}
\sum_{t=0}^\infty \eta_t = \infty, \quad \sum_{t=0}^\infty \eta_t^2 < \infty, \quad \sum_{t=0}^\infty \rho_t = \infty, \quad \sum_{t=0}^\infty \rho_t^2 < \infty, \quad \lim_{t\to\infty} \frac{\rho_t}{\eta_t} = 0.
\end{align*}

\item \emph{Martingale difference noise}: $\xi^{(Q)}_t$ and $\xi^{(\mu)}_t$ are martingale difference sequences with respect to the filtration $\mathcal{F}_t$, information available up to time 
$t$ (random variable with zero conditional mean) :
\begin{align*}
\mathbb{E}[\xi^{(Q)}_t | \mathcal{F}_t] = 0, \quad \mathbb{E}[\xi^{(\mu)}_t | \mathcal{F}_t] = 0, \quad \mathbb{E}[(\xi^{(Q)}_t)^2 | \mathcal{F}_t] \leq K, \quad \mathbb{E}[(\xi^{(\mu)}_t)^2 | \mathcal{F}_t] \leq K.
\end{align*}

\item \emph{Bounded iterates}: The sequences $\{Q_t\}$, $\{\mu_t\}$, $\{M_t\}$ remain bounded almost surely.

\item \emph{Memory attractor}: The memory process $\{M_t\}$ has a compact attractor set $ M_{\infty} \subset  M$. This means memory size is bounded and the content distribution stabilises over time.
In practice, this is achieved by: 1) Using fixed-size memory, 2)
Having policies that converge to stationary distributions, 3)
Using environments that produce stationary experience streams, or
4) Conservative memory update strategies.
\end{enumerate}
\end{assumption}

\begin{theorem}[Two-Time-Scale Convergence for Parzen-KL Policy Iteration]
\label{thm:twots-parzen-detailed}
Under Assumptions~\ref{as:bounded} and \ref{as:twots-detailed}, let $(Q_t,\mu_t,M_t)$ be generated by the following \emph{sample-based} recursions.

\begin{enumerate}
\item \emph{Q-update (TD learning, sample form)}:
\begin{align}
Q_{t+1}(x_t,c_t)
&= Q_t(x_t,c_t) + \eta_t\Big[
\delta_t(Q_t,\mu_0,M_{t+1})
\Big],
\label{eq:q-update-sample}
\\[-0.2em]
\delta_t(Q,\mu_0,M)
&:= r_t + \gamma \alpha \log\!\Big(
\sum_{c' \in M} \mu_0(c'|x_{t+1}) e^{Q(x_{t+1},c')/\alpha}
\Big) - Q(x_t,c_t).
\nonumber
\end{align}

\item \emph{Policy update (Parzen-KL improvement, sample form)}:
\begin{align}
\mu_{t+1}(c|x_t)
&= \mu_t(c|x_t) + \eta_t\Big[
\Phi_t(c|x_t;Q_t,\mu_0,M_t) - \mu_t(c|x_t)
\Big],
\label{eq:policy-update-sample}
\\[-0.2em]
\Phi_t(c|x;Q,\mu_0,M)
&:= \frac{\mu_0(c|x)\,e^{Q(x,c)/\alpha}}{Z_t(x)},
\qquad
Z_t(x) := \sum_{c' \in M} \mu_0(c'|x)\,e^{Q(x,c')/\alpha}.
\nonumber
\end{align}

\item \emph{Memory update (experience replay)}:
\begin{align}
M_{t+1} =
\begin{cases}
M_t \cup \{(s_t,a_t,r_t)\} & \text{with probability } \rho_t,\\
M_t & \text{otherwise.}
\end{cases}
\label{eq:memory-update-detailed}
\end{align}
\end{enumerate}

Define the filtration $\mathcal{F}_t := \sigma\big((x_k,c_k,r_k), M_k, Q_k, \mu_k : k \le t\big)$, which collects all information available to the agent up to time (t), including past observations, the current memory state, and the value function and policy parameters. Define the \emph{mean drifts} (expected operators) by
\begin{align}
H_Q(Q_t,\mu_t,M_t)(x_t,c_t)
&:= \mathbb{E}\!\left[\,\delta_t(Q_t,\mu_0,M_{t+1}) \mid \mathcal{F}_t\right],
\label{eq:HQ-def}
\\
H_\mu(Q_t,\mu_t,M_t)(c|x_t)
&:= \mathbb{E}\!\left[\,\Phi_t(c|x_t;Q_t,\mu_0,M_t) - \mu_t(c|x_t)\mid \mathcal{F}_t\right].
\label{eq:Hmu-def}
\end{align}
Then define the \emph{martingale difference noises} as the residuals
\begin{align}
\xi^{(Q)}_{t+1}
&:= \delta_t(Q_t,\mu_0,M_{t+1}) - H_Q(Q_t,\mu_t,M_t)(x_t,c_t),
\label{eq:xiQ-def}
\\
\xi^{(\mu)}_{t+1}(c|x_t)
&:= \Big(\Phi_t(c|x_t;Q_t,\mu_0,M_t) - \mu_t(c|x_t)\Big) - H_\mu(Q_t,\mu_t,M_t)(c|x_t).
\label{eq:xiu-def}
\end{align}
With these definitions, the recursions can be written in stochastic approximation form:
\begin{align}
Q_{t+1}(x_t,c_t) &= Q_t(x_t,c_t) + \eta_t\Big[ H_Q(Q_t,\mu_t,M_t)(x_t,c_t) + \xi^{(Q)}_{t+1}\Big],
\label{eq:q-update-SA}
\\
\mu_{t+1}(c|x_t) &= \mu_t(c|x_t) + \eta_t\Big[ H_\mu(Q_t,\mu_t,M_t)(c|x_t) + \xi^{(\mu)}_{t+1}(c|x_t)\Big].
\label{eq:policy-update-SA}
\end{align}

Moreover, $(\xi^{(Q)}_{t+1})$ and $(\xi^{(\mu)}_{t+1})$ are martingale difference sequences w.r.t.~$(\mathcal{F}_t)$:
\[
\mathbb{E}\!\left[\xi^{(Q)}_{t+1}\mid\mathcal{F}_t\right]=0,\qquad
\mathbb{E}\!\left[\xi^{(\mu)}_{t+1}(c|x_t)\mid\mathcal{F}_t\right]=0,
\]
and satisfy the conditional second-moment bounds in Assumption~\ref{as:twots-detailed} ii.

Finally, the conclusions (a)--(c) hold:
\begin{enumerate}[label=(\alph*)]
\item The fast variables $(Q_t,\mu_t)$ track the stationary-memory soft policy-iteration fixed points:
\[
\limsup_{t\to\infty}\inf_{(Q^*_M,\mu^*_M)\in\mathcal{F}(M_t)}
\|(Q_t,\mu_t)-(Q^*_M,\mu^*_M)\|\le \epsilon\quad\text{a.s.}
\]
\item The joint process converges to the limiting set:
\[
(Q_t,\mu_t,M_t)\to\{(Q^*_M,\mu^*_M,M): M\in M_\infty,\ (Q^*_M,\mu^*_M)\in\mathcal{F}(M)\}.
\]
\item If $M_t\to M_\infty$ a.s., then $(Q_t,\mu_t)\to(Q^*_{M_\infty},\mu^*_{M_\infty})$.
\end{enumerate}
\end{theorem}


Appendix~\ref{Lipschitz Properties of Memory Updates} presents the required Lipschitz properties of the memory updates, and Appendix~\ref{proof:two-time-scale} provides the detailed proof of Theorem~\ref{thm:twots-parzen-detailed}.
The novelty lies in extending the classical ODE method to handle structured, state-dependent episodic memory updates and their coupling with fast policy dynamics under a two–time-scale regime. The key insight is that from the perspective of the fast variables $(Q_t, \mu_t)$, the slow variable $M_t$ appears quasi-static due to the timescale separation $\rho_t/\eta_t \to 0$. This detailed proof establishes several key properties of our \emph{Read--Write Learning} iteration. First, it demonstrates robustness to memory updates, showing that the algorithm remains stable even as the memory evolves, provided that memory updates occur on a slower timescale than value and policy updates. Second, it guarantees tracking behaviour, ensuring that the value function and policy rapidly adjust to reflect changes in memory, remaining close to the optimal configuration for the current memory state. Third, the analysis confirms convergence to local optima, proving that for each fixed memory configuration (M), the algorithm converges to the corresponding optimal retrieval policy ($\mu^*_M$). Finally, the timescale separation result provides practical implementation guidance, indicating that memory should be updated less frequently than the value function and policy.

\begin{algorithm}[t]
\caption{Read--Write Reflective Learning (Fast Policy Update, Slow Memory Write)}
\label{alg:Read--Write Reflective Learning}
\begin{algorithmic}[1]
\State \textbf{Input:} Temperature $\alpha>0$, kernel bandwidth $h>0$, discount $\gamma\in(0,1)$
\State \textbf{Stepsizes:} $\{\eta_t\}$ for policy evaluation/improvement, $\{\rho_t\}$ for memory updates,
with $\rho_t/\eta_t \to 0$, $\sum_{t=1}^\infty \eta_t=\infty$, $\sum_{t=1}^\infty \eta_t^2<\infty$
\State \textbf{Initialise:} Memory $M_0$, Q-function $Q_0$, initial Parzen prior $\mu_{0,0}(\cdot\mid x)$ (incl.\ void case $c_{\varnothing}$)
\For{$t=0,1,2,\ldots$}
\State \textbf{(1) State and Prior Construction}
\State Observe $s_t$ and form augmented state $x_t=(s_t,M_t)$.
\State Construct the reference prior $\mu_{0,t}(\cdot\mid x_t)$ from $M_t$ using~\eqref{eq:parzen-prior} and~\eqref{eq:mu0-mixture}.
\State \textbf{(2) Read: Policy Improvement (Fast Time Scale)}
\State Define the KL-greedy retrieval policy (Eq.~\ref{eq:parzen-kl-solution}) induced by $Q_t$:
\[
\mu_t(c\mid x_t)
=
\frac{\mu_{0,t}(c \mid x_t)\exp(Q_t(x_t,c)/\alpha)}
{\sum_{c'}\mu_{0,t}(c' \mid x_t)\exp(Q_t(x_t,c')/\alpha)}.
\]
\State Sample a retrieval action $c_t \sim \mu_t(\cdot\mid x_t)$.
\State \textbf{(3) LLM Act}
\State Sample an environment action $a_t \sim p_{\mathrm{LLM}}(\cdot\mid s_t,c_t)$ and execute it.
\State \textbf{(4) Environment Feedback}
\State Observe reward/feedback $r_t$ and next state $s_{t+1}$ 

   \State \textbf{(5) a. Write (Slow Memory Update):} update episodic memory,
    \[
    M_{t+1} \leftarrow (1-\rho_t)M_t + \rho_t\,\mathrm{Write}(M_t,s_t,a_t,r_t,s_{t+1}) \,,
    \]
    which denotes a generic slow update consistent with the assumed Lipschitz conditions.
     \State Next augmented state $x_{t+1} := (s_{t+1},M_{t+1})$; Next prior $\mu_{0,t+1}(\cdot\mid x_{t+1})$ from $M_{t+1}$
     \State \textbf{\ \ \ \ \ b. Write (Fast KL-regularised Evaluation):} compute the target
    \[
    y_t := r_t + \gamma\Bigg(
      \sum_{c'\in \mathcal C(M_{t+1})}\mu_t(c'\mid x_{t+1})\,Q_t(x_{t+1},c')
      - \alpha\,\mathrm{KL}\!\big(\mu_t(\cdot\mid x_{t+1})\;\|\;\mu_{0,t+1}(\cdot\mid x_{t+1})\big)
    \Bigg),
    \]
    and update
    \[
    Q_{t+1}(x_t,c_t) \leftarrow (1-\eta_t)Q_t(x_t,c_t) + \eta_t\,y_t,
    \]
    which is stored in memory in the tabular case, or updated via function approximation.

\EndFor
\end{algorithmic}
\end{algorithm}

\subsection{The Role of Reflection and Memory towards Optimal Policy}
We are now ready to present the complete \emph{Read--Write Reflective Learning} pseudocode in Algorithm~\ref{alg:Read--Write Reflective Learning}, consistent with the workflow illustrated in Fig.~\ref{fig:cbr}.

To ensure practical feasibility, policy evaluation is implemented in an approximate form using feedback from the environment. Steps (2)–(4) and (5.b) constitute a fast policy iteration loop. The Read step performs a KL-regularised greedy update of the retrieval policy, while Step (5.b) approximately evaluates the resulting policy using feedback from the environment.
The policy improvement is monotonic because the Read step solves a KL-regularised optimisation problem with respect to the current value function (Lemma \ref{lemma-objective}). Given a fixed reference prior, the resulting update maximises the regularised expected return and therefore yields a policy that is no worse than the previous one in terms of the KL-regularised objective. Since the subsequent evaluation step provides an unbiased (or consistent) estimate of the policy value under environmental feedback, repeated application of Steps (2)–(4) and (5.b) leads to monotonic improvement of the retrieval policy on the fast time scale (Theorem \ref{thm:parzen-improvement}). Step (5.a) implements the slow evolution of episodic memory. Under the two-time-scale assumptions, Theorem~\ref{thm:twots-parzen-detailed} establishes that the resulting algorithm converges to the optimal policy of the reflected MDP.

Crucially, episodic memory does not merely augment the state representation but
directly parameterises the effective Bellman operator. Each memory update induces
a new Reflected MDP and hence a new policy-evaluation problem. Although the
$Q$-update takes the form of a standard temporal-difference step, the associated
KL-evaluation operator is indexed by the current memory $M$ through the induced
action set $\mathcal{C}(M)$, the memory-conditioned prior $\mu_0(\cdot \mid x, M)$,
and the memory-dependent transition $x' = (s', \mathrm{Write}(M,\cdot))$.

Consequently, the read--write cycle realises a two-time-scale policy-iteration
procedure: for fixed $M$, the effective MDP is stationary and the algorithm
converges to the locally optimal policy induced by the current memory coverage,
while memory growth drives successive improvements of the effective control
problem. Since the Reflected MDP approximates the underlying environment MDP
through LLM-mediated interaction (cf.~\eqref{three-mdps}), a central question is
whether a Read--Write Reflective Learning agent can asymptotically achieve optimal
MDP performance as its episodic memory expands. We formalise this requirement via
\emph{reflection consistency}, which states that increasing memory coverage drives
the composite policy—combining retrieval and LLM action selection—toward the true
optimal policy.

\begin{figure}[t]
    \centering
    \begin{tikzpicture}[x=1cm,y=1cm,>=Latex]
  \tikzstyle{case}=[circle, draw, line width=0.6pt, minimum size=4mm, inner sep=0pt]
  \definecolor{goodc}{RGB}{47,158,68}
  \definecolor{badc}{RGB}{180,50,50}
  \definecolor{llmcol}{RGB}{66,135,245}
  \definecolor{optcol}{RGB}{245,150,60}
  \definecolor{tvshade}{RGB}{210,210,210}

  \begin{scope}
    \coordinate (s) at (0,0);
    \draw[thick] (s) node[below left=-1pt] {$s$} circle [radius=1.4];
    \node[fill=black,circle,inner sep=1.2pt,label={below:$s$}] at (s) {};

    \draw[gray,->] (s) -- ++(1.1,0) node[midway,below] {$r$};

    \node[case,draw=goodc] (cgood) at (0.9,0.45) {};
    \node[above right=-1pt of cgood,scale=0.8,text=goodc] {\(\text{good }c\)};
    \node[case,draw=badc] (cbad1) at (-0.4,0.9) {};
    \node[case,draw=badc] (cbad2) at (-1.3,-0.2) {};
    \node[case,draw=badc] (cout) at (2.2,0.3) {}; 

    \node[gray,scale=0.8] at (0,1.55) {$B(s,r)$};

    \draw[rounded corners,gray] (-1.8,-1.4) rectangle (2.6,1.6);
    \node[gray,scale=0.8] at (1.95,1.35) {episodic memory};

    \draw[->,thick,goodc] (cgood) to[bend left=15] (3.6,0.8)
      node[pos=0.4,above,scale=0.85,text=goodc] {retrieve \(c\)};

    \node[gray,scale=0.8,align=center] at (-1.0,-1.1)
      {$d\big(s,s(c)\big)\le r$};
  \end{scope}

  \begin{scope}[xshift=4.5cm,yshift=-1.2cm]
    \draw[->] (-0.2,0) -- (5.5,0) node[right] {actions};
    \draw[->] (0,-0.2) -- (0,2.8) node[above] {probability};

    \foreach \i/\name in {0.5/$a_1$,1.5/$a_2$,2.5/$a_3$,3.5/$a_4$,4.5/$a_5$} {
      \draw (\i,0) -- (\i,-0.08) node[below,yshift=-1pt] {\name};
    }

    \def\pA{1.9} \def\pB{1.1} \def\pC{0.6} \def\pD{0.9} \def\pE{0.5}
    \def\qA{1.6} \def\qB{1.2} \def\qC{0.8} \def\qD{0.7} \def\qE{0.4}

    \foreach \x/\h in {0.5/\pA,1.5/\pB,2.5/\pC,3.5/\pD,4.5/\pE} {
      \draw[fill=llmcol,opacity=0.35,draw=llmcol] (\x-0.28,0) rectangle (\x-0.02,\h);
    }

    \foreach \x/\h in {0.5/\qA,1.5/\qB,2.5/\qC,3.5/\qD,4.5/\qE} {
      \draw[fill=optcol,opacity=0.35,draw=optcol] (\x+0.02,0) rectangle (\x+0.28,\h);
    }

    \foreach \x/\p/\q in {0.5/\pA/\qA,1.5/\pB/\qB,2.5/\pC/\qC,3.5/\pD/\qD,4.5/\pE/\qE} {
      \pgfmathsetmacro{\top}{max(\p,\q)}
      \pgfmathsetmacro{\bot}{min(\p,\q)}
      \draw[fill=tvshade,opacity=0.5,draw=none] (\x-0.28,\bot) rectangle (\x+0.28,\top);
    }

    \draw[fill=llmcol,opacity=0.35,draw=llmcol] (0.2,2.35) rectangle ++(0.3,0.18);
    \node[anchor=west] at (0.55,2.44) {$p_{\mathrm{LLM}}(\cdot\mid s,c)$};
    \draw[fill=optcol,opacity=0.35,draw=optcol] (3.2,2.35) rectangle ++(0.3,0.18);
    \node[anchor=west] at (3.55,2.44) {$\pi^\star(\cdot\mid s)$};
    \draw[fill=tvshade,opacity=0.5,draw=black!60] (5.2,2.35) rectangle ++(0.3,0.18);
    \node[anchor=west] at (5.55,2.44) {\(\tfrac{1}{2}\sum_a |p-q|\) (TV)};

    \draw[<->,gray] (4.9,0.4) -- (4.9,1.3);
    \node[gray,anchor=west,scale=0.85] at (5.0,0.85) {TV contribution};
  \end{scope}
\end{tikzpicture}\caption{The illustration of the assumption of LLM local sufficiency.}
    \label{fig:placeholder}
\end{figure}
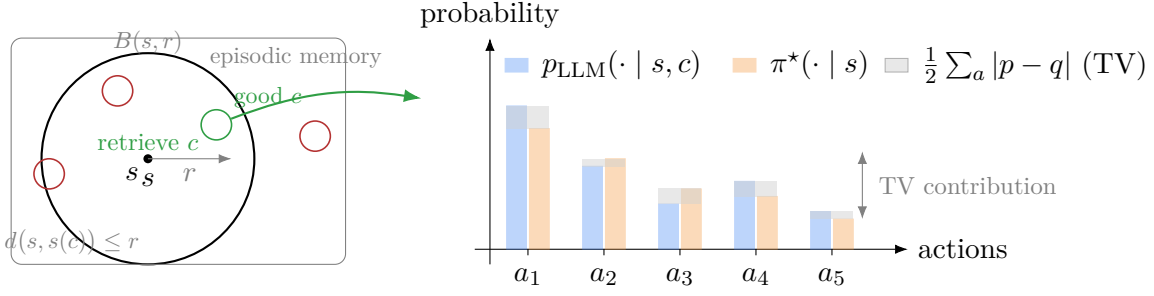

For an episodic memory $M \subset \mathfrak{M}$, define the coverage radius
\[
r_M := \sup_{s \in \mathcal S} \min_{c \in M} d\bigl(s, s(c)\bigr),
\]
where $s(c)$ denotes the state component of memory case $c$.  
Equivalently, the statewise coverage radius is $r_M(s) := \min_{c \in M} d\bigl(s, s(c)\bigr)$.

Let the (deterministic) retrieval rule $c_M(s)$ pick any case achieving the minimum.
The LLM-induced composite policy on the \emph{original} state space is
\[
\pi_M(a\mid s) \;:=\; p_{\mathrm{LLM}}\big(a \mid s, c_M(s)\big).
\]
Assume bounded rewards $| \mathcal{R}(s,a) | \le R_{\max}$ and $\gamma\in[0,1)$.

\begin{assumption}[LLM local consistency]\label{as:llm-consistency}
There exists a modulus $\varepsilon_{\mathrm{LLM}}(r)\rightarrow 0$ as $r\rightarrow 0$ such that
for any $s\in\mathcal{S}$ and any case $c$ with $d\big(s,s(c)\big)\le r$,
\[
\mathrm{TV}\!\left(p_{\mathrm{LLM}}(\cdot\mid s,c),\, \pi^*(\cdot\mid s)\right) \;\le\; \varepsilon_{\mathrm{LLM}}(r),
\]
where $\mathrm{TV}(p,q)$ denotes the total variation distance:
\[
\mathrm{TV}(p,q) = \tfrac{1}{2} \sum_{a \in \mathcal{A}} \left| p(a) - q(a) \right| 
\quad \text{(discrete case)},
\]
or
\[
\mathrm{TV}(p,q) = \tfrac{1}{2} \int_{\mathcal{A}} \left| p(a) - q(a) \right| \, da
\quad \text{(continuous case)}.
\]\end{assumption}

The function $\varepsilon_{\mathrm{LLM}}(r)$ characterises the local decision-making capability
of the LLM and can be interpreted as a measure of its \emph{local optimality gap}.
Specifically, it quantifies how closely the LLM-induced action distribution
$p_{\mathrm{LLM}}(\cdot \mid s, c)$ approximates the optimal policy $\pi^\star(\cdot \mid s)$
when the retrieved memory case $c$ lies within a neighbourhood of radius $r$ around the true
state $s$.

A small value of
$\varepsilon_{\mathrm{LLM}}(r)$ indicates that the LLM is able to infer near-optimal actions
using only coarse or approximate contextual information. In other words, even when the retrieved
memory does not exactly match the current state, a strong LLM can leverage its internal semantic
representations, world knowledge, and reasoning ability to interpolate correctly within a
local region of the state space. Consequently, a \emph{good} LLM is characterised by an error
function $\varepsilon_{\mathrm{LLM}}(r)$ that remains small even for relatively large values
of $r$, reflecting robustness to imprecise memory retrieval and limited state coverage.

Intuitively, $\varepsilon_{\mathrm{LLM}}(r)$ captures the effective ``radius of competence''
of the LLM: it measures the size of the neighbourhood within which the LLM can reliably recover
near-optimal behaviour from local contextual cues. This property directly determines how densely
episodic memory must cover the state space in order to achieve near-optimal performance. A smaller
$\varepsilon_{\mathrm{LLM}}(r)$ for larger $r$ implies that fewer memory cases are required,
leading to improved sample efficiency and faster convergence of the Read--Write Reflective
Learning process.

We now examine the retrieval capability. For a given state $s$, define $\delta_M(s)$ as the probability
that the retriever selects a case whose associated state lies outside radius $r_M(s)$, conditioned on
the existence of at least one case within this radius, due to auxiliary retrieval errors (e.g.,
finite-bandwidth smoothing or finite-sample density estimation):
\[
\delta_M(s)
\;:=\;
\mathbb{P}\!\left(
d\!\left(s, s(c)\right) > r_M(s)
\;\middle|\;
\exists\, c' \in M :
d\!\left(s, s(c')\right) \le r_M(s)
\right).
\]
\begin{lemma}[Retrieval Capability]
\label{lem:policy-approximation}
Under Assumption \ref{as:llm-consistency}:
$$\mathrm{TV}(\pi_M(\cdot\mid s), \pi^\star(\cdot\mid s)) \leq \varepsilon_{\mathrm{LLM}}(r_M(s)) + \delta_M(s)$$.
\end{lemma}
\begin{proof}
Let $\mathcal{E}_s$ denote the event that the retriever selects a \emph{good} case within
distance $r_M(s)$ of state $s$. By construction, this event occurs with probability at
least $1-\delta_M(s)$. Conditioned on $\mathcal{E}_s$, Assumption~\ref{as:llm-consistency} implies
\[
\mathrm{TV}\!\left(\pi_M(\cdot \mid s), \pi^\star(\cdot \mid s)\right)
\le
\varepsilon_{\mathrm{LLM}}\!\left(r_M(s)\right).
\]

With the complementary event $\mathcal{E}_s^{c}$, the total variation distance
is trivially bounded by
\[
\mathrm{TV}\!\left(\pi_M(\cdot \mid s), \pi^\star(\cdot \mid s)\right) \le 1.
\]

Taking expectation over the retrieval event yields
\begin{align*}
\mathrm{TV}\!\left(\pi_M(\cdot \mid s), \pi^\star(\cdot \mid s)\right)
&\le
(1-\delta_M(s))\,\varepsilon_{\mathrm{LLM}}\!\left(r_M(s)\right)
+
\delta_M(s) \\
&\le
\varepsilon_{\mathrm{LLM}}\!\left(r_M(s)\right)
+
\delta_M(s),
\end{align*}
which completes the proof.
\end{proof}

\begin{lemma}[Immediate reward approximation]\label{lem:reward-approx}
For any $s\in\mathcal{S}$, with $c^*=c_M(s)$,
\[
\bigg|\; \mathcal{R}_{\mathrm{LLM}}\big((s,M),c^*\big) \;-\; \sum_{a}\pi^*(a\mid s)\,\mathcal{R}(s,a) \;\bigg|
\;\le\; 2 R_{\max}\, \mathrm{TV}\!\Big(p_{\mathrm{LLM}}(\cdot\mid s,c^*),\,\pi^*(\cdot\mid s)\Big).
\]
\end{lemma}
\begin{proof}[Proof]
Write $\mathcal{R}_{\mathrm{LLM}}\!=\!\sum_a p_{\mathrm{LLM}}(a\mid s,c^*)\,\mathcal{R}(s,a)$ and
$R^*(s)\!=\!\sum_a \pi^*(a\mid s)\,\mathcal{R}(s,a)$.
By Hölder’s inequality and the definition of total variation,
$\big|\sum_a (p-\pi^*)\mathcal{R}\big|\le \|\mathcal{R}\|_\infty \|p-\pi^*\|_1
\le 2R_{\max}\,\mathrm{TV}(p,\pi^*)$.
\end{proof}

\begin{theorem}[Value difference bound via policy TV]\label{thm:value-tv}
Let $\pi^*$ be the optimal for the original MDP, and let $\pi_M$ be the composite policy defined in SRDP.
Then
\[
\big\| V^{\pi^*} - V^{\pi_M} \big\|_{\infty}
\;\le\;
\frac{2 R_{\max}}{(1-\gamma)^2}\;
\sup_{s\in\mathcal{S}} \mathrm{TV}\!\big(\pi_M(\cdot\mid s),\,\pi^*(\cdot\mid s)\big)
\;=\;
\frac{2 R_{\max}}{(1-\gamma)^2}\, \Delta_M.
\]
\end{theorem}
\begin{proof}[Proof]
Consider the Bellman operators $(T^{\pi}V)(s)=\sum_a \pi(a\mid s)\big(\mathcal{R}(s,a)+\gamma\,\mathbb{E}[V(s')\mid s,a]\big)$.
Since $V^{\pi^*}$ satisfies $V^{\pi^*}=T^{\pi^*}V^{\pi^*}$,
\[
\| V^{\pi^*} - V^{\pi_M} \|_\infty
\;\le\; \frac{1}{1-\gamma} \,\big\| (T^{\pi^*}-T^{\pi_M}) V^{\pi^*} \big\|_\infty.
\]
For any $s$,
\begin{align*}
&\big|(T^{\pi^*}-T^{\pi_M}) V^{\pi^*}(s)\big| \\
&= \Big| \sum_a \big(\pi^*-\pi_M\big)(a\mid s)\,\mathcal{R}(s,a)
\;+\; \gamma\, \sum_a \big(\pi^*-\pi_M\big)(a\mid s)\,\mathbb{E}[V^{\pi^*}(s')\mid s,a] \Big| \\
&\le 2R_{\max}\,\mathrm{TV}(\pi^*,\pi_M) \;+\; \gamma \,\|V^{\pi^*}\|_\infty \cdot 2\,\mathrm{TV}(\pi^*,\pi_M),
\end{align*}
using Lemma~\ref{lem:reward-approx}’s argument and that $|\mathbb{E}V^{\pi^*}|\le \|V^{\pi^*}\|_\infty$.
Since $\|V^{\pi^*}\|_\infty \le R_{\max}/(1-\gamma)$,
\[
\big\|(T^{\pi^*}-T^{\pi_M}) V^{\pi^*}\big\|_\infty
\;\le\; 2R_{\max}\,\mathrm{TV}(\pi^*,\pi_M)\Big(1+\tfrac{\gamma}{1-\gamma}\Big)
\;=\; \frac{2R_{\max}}{1-\gamma}\,\mathrm{TV}(\pi^*,\pi_M).
\]
Divide by $(1-\gamma)$ to get the claimed bound.
\end{proof}

\begin{corollary}[Asymptotic optimality with growing memory]
\label{cor:asymp-opt-parzen}
Under Assumption~\ref{as:llm-consistency} and Lemma~\ref{lem:policy-approximation}, define 
\[
\delta_M \;:=\; \sup_{s\in\mathcal{S}} \delta_M(s).
\]
Then
\[
\Delta_M
\;:=\;
\sup_{s\in\mathcal{S}}
\mathrm{TV}\!\big(\pi_M(\cdot\mid s), \pi^\star(\cdot\mid s)\big)
\;\le\;
\varepsilon_{\mathrm{LLM}}(r_M) + \delta_M.
\]
If $r_M \to 0$ and $\delta_M \to 0$, then $\Delta_M \to 0$ and
\[
\sup_{s\in\mathcal{S}} \big| V^{\pi^\star}(s) - V^{\pi_M}(s) \big|
\;\le\;
\frac{2 R_{\max}}{(1-\gamma)^2}\,\Delta_M
\;\longrightarrow\; 0.
\]
Hence, as memory grows, the Read--Write operation drives the agent toward optimal
performance in the underlying MDP.
\end{corollary}

\begin{proof}
Lemma~\ref{lem:policy-approximation} implies
$\mathrm{TV}(\pi_M(\cdot\mid s), \pi^\star(\cdot\mid s))
\le \varepsilon_{\mathrm{LLM}}(r_M) + \delta_M(s)$
for all $s\in\mathcal{S}$. Taking the supremum over $s$ yields
$\Delta_M \le \varepsilon_{\mathrm{LLM}}(r_M) + \delta_M \to 0$.
Applying Theorem~\ref{thm:value-tv} completes the proof.
\end{proof}

Corollary~15 formalises the role of episodic memory in enabling asymptotically optimal
behaviour without parameter updates. The result shows that the discrepancy between the
memory-induced policy and the optimal policy of the underlying MDP decomposes into an
LLM approximation error and a memory coverage error. As episodic memory grows and achieves sufficient coverage of the state space, in the sense that locally relevant memory cases enable the LLM to approximate optimal actions within neighbourhoods of each state, both error terms vanish. Consequently, the composite policy and its associated value function converge to the optimal solution. This establishes that
iterative read--write reflection over expanding memory is sufficient to recover optimal
performance in the limit.

\section{Related Work}

Our work on reflective memory dynamics and Read-Write Reflective Learning builds upon, yet also departs from, several established lines of research in reinforcement learning and large language models (LLMs). This section situates our contribution within the broader landscape of memory-based RL and clarifies the key distinctions from existing approaches.

The integration of memory mechanisms into reinforcement learning has long been explored to address the challenges of partial observability and sample complexity. Early approaches often relied on recurrent neural networks (RNNs), in particular Long Short-Term Memory (LSTM) networks \citep{hochreiter1997long}, to maintain hidden states that capture historical information. For instance, \citep{hausknecht2015deep} extends DQN with an LSTM to summarise past observations for Q-learning under partial observability, providing an early neural-memory approach to POMDPs \cite{kaelbling1998planning}. However, its memory is implicit and entangled in the recurrent hidden state, making long-horizon retention, interpretability, and revision of past information difficult. 

More recent methods have taken inspiration from cognitive neuroscience, especially the hippocampal processes of memory formation and retrieval in humans and animals. For instance, the Hippocampal-Augmented Memory Integration (HAMI) framework \citep{hami2025} incorporates symbolic indexing and hierarchical refinement based on hippocampal functions. Although HAMI shares our biological motivation, our Parzen-KL retrieval mechanism diverges in its probabilistic formulation and information-theoretic regularisation, which underpin both stability and scalability.

From another strand of memory-based RL, episodic control methods \citep{ramani2019short} such as Model Free Episodic Control  \citep{blundell2016model} and Neural Episodic Control \citep{pritzel2017neural} use memory primarily as a direct control mechanism, retrieving past transitions to approximate values or reuse actions via similarity matching. In this setting, memory effectively acts as a nonparametric policy, and generalisation is limited to local interpolation over stored experiences \citep{ramani2019short}.

In contrast, our \emph{Read--Write Reflective Learning} uses episodic memory to enable reflection in large language models. Retrieved memory does not prescribe actions directly; instead, it provides local contextual grounding that allows the LLM to reason and generate actions within a semantically relevant neighbourhood. Memory thus serves as a substrate for reflective generalisation rather than as a policy itself. By combining density based similarity priors with KL regularised updates, our approach yields a principled, scalable retrieval mechanism that supports stable continual learning while exploiting the LLM’s inherent capacity for local semantic generalisation.

Recent work on Stable Hadamard Memory \citep{hadamard2024} highlights the importance of stability in memory systems, using Hadamard products to calibrate and update stored information. While complementary in spirit, our method focuses instead on retrieval optimisation through information-theoretic regularisation. These two perspectives, stability of storage and optimality of retrieval, could be integrated in future research to build more comprehensive memory architectures.

Case-based reasoning (CBR) \citep{AamodtPlaza1994,Ashley1990} has long inspired memory-driven problem solving through the retrieval and adaptation of past cases. Our prior work on CBR-based LLMs \cite{guo2024ds,guo2025optimizing} provides empirical evidence, particularly in data science and coding tasks, that LLMs are highly effective at revising retrieved cases, enabling strong generalisation to new tasks and overcoming a critical bottleneck in traditional CBR systems. Within our framework, CBR-based LLMs can be viewed as a stateless special case, for which we provide a principled and unified treatment.

Experience replay, a widely used mechanism in deep RL since its adoption in DQN \citep{mnih2015human}, has evolved from uniform sampling to prioritised replay \citep{schaul2015prioritized}, where transitions are sampled according to temporal-difference errors. While these methods improve sample efficiency, they do not provide the structured, similarity-based retrieval required for generalisation across diverse contexts. Our approach diverges here by employing density-aware retrieval via Parzen window estimation, offering more nuanced and context-sensitive access to stored experiences.

Our work also connects to broader developments in information-theoretic reinforcement learning. Maximum entropy RL \citep{haarnoja2018soft} introduces entropy regularisation to encourage exploration and robustness, while KL constraints in algorithms such as TRPO \citep{schulman2015trust} and MPO \citep{abdolmaleki2018maximum} ensure stable policy updates. The distinctive aspect of our contribution is the application of KL regularisation not only to the policy but also to the retrieval process itself. This dual-level regularisation balances the exploitation of high-value memories with the exploration of diverse experiences, whilst preserving consistency with the underlying similarity structure.

Finally, from a theoretical perspective, our framework extends established convergence results in reinforcement learning. The convergence proof for our soft policy iteration builds upon entropy-regularised MDPs \citep{neu2017unified} but adapts them to the context of memory-based retrieval. Our two-time-scale analysis further draws on stochastic approximation theory \citep{borkar2009stochastic}, extending it to SRDPs where value and memory are updated simultaneously. A central outcome is the formal equivalence we establish between read (retrieval) and write operations and the fundamental processes of policy improvement and policy evaluation. This connection makes retrieval-augmented LLMs provably learnable for the first time, opening the way for reinforcement learning theory and tools to be systematically applied in this emerging setting.

\section{Conclusions and Future Work}
In this paper, we have established a formal theoretical foundation for continual learning in LLM agents through episodic memory and reflection, without requiring parameter fine-tuning. The core contribution is the \emph{Stateful Reflective Decision Process (SRDP)}, which extends the classical MDP framework by modeling reflection as a dual-action process: a retrieval stage that reads from episodic memory (policy improvement) and an action generation stage conditioned on retrieved cases. By augmenting the state space to include memory, we prove this system is Markovian and can be analysed as a \emph{Reflected MDP}. We then introduce \emph{Read-Write Reflective Learning}, a practical algorithm that integrates Parzen-window-based retrieval with KL-regularised soft policy iteration. Key theoretical results include: (1) guaranteed convergence of the policy iteration algorithm under bounded rewards, (2) two-timescale convergence when memory evolves slowly relative to policy updates, and (3) asymptotic optimality: as episodic memory grows to cover the state space, the composite policy provably converges to the optimal policy of the underlying MDP. This framework unifies case-based reasoning \cite{guo2024ds}, retrieval-augmented generation, and reinforcement learning under a single mathematical formalism, providing the first rigorous convergence guarantees for memory-driven LLM agents.

Beyond the theoretical advances, the analysis highlights several practical implications for the design of memory-augmented LLM agents. Effective performance requires density-aware retrieval strategies to balance exploration and exploitation.  The void case $c_0$ controls the discovery from the LLM itself whose parameter is required to be tuned. Stability can be enhanced by implementing soft, entropy-regularised improvements, while systematic memory growth helps expand state coverage. In addition, similarity metrics and bandwidths should adapt as the memory evolves to maintain efficiency and relevance.

Despite these strengths, the framework has certain limitations. It might face computational challenges as memory size increases. Moreover, the impact of embedding quality on performance is not fully addressed.

Looking ahead, several promising directions for future work emerge. In particular, we plan to internalise episodic memory within the LLM architecture itself, yielding a unified stateful neural machine, which we outline next.

\subsection{Internalising Episodic Memory into the LLM Architecture}
In modern contexts, LLM agents increasingly employ explicit episodic memory systems to extend context and improve performance across diverse tasks \citep{Zhong2024,Chhikara2025,fountas2024human}. Yet, the integration of memory retrieval with decision-making in LLMs still lacks a firm theoretical foundation, particularly regarding the conditions under which such systems converge to optimal behaviour. Our work addresses this gap directly.

A natural next step is to integrate the episodic memory and read--write reflection mechanism of SRDP directly into the LLM architecture, yielding a single neural system whose internal state
corresponds to the augmented state $(s,M)$. Recent ``test-time
memorisation'' lines of work provide useful architectural blueprints. In particular,
Titans introduces a neural long-term memory module that updates while the model is running,
thereby supporting continual memorisation beyond the context window without offline retraining
\cite{behrouz2024titans}. MIRAS further reframes modern sequence models as instances of
associative memory with explicit choices of memory structure, retention, and online
optimisation \cite{behrouz2025miras}. Our EM-LLM work in~\citep{fountas2024human} introduces a human-inspired episodic memory architecture that enables large language models to operate with effectively infinite context by selectively storing, retrieving, and reusing past experiences to support long-horizon reasoning and decision-making.
These developments suggest a promising direction: replace external retrieval of SRDP over a growing case base by an internal memory state that is updated online, while retaining the SRDP control view in which ``read'' implements policy improvement and ``write'' implements policy evaluation.

Our Read-Write loop differs from Titans-style test-time updates in a key way: SRDP performs
stateful adaptation through \emph{experience-grounded reflection} and memory evolution, rather
than gradient-based parameter updates of the base model. A concrete research direction is to
design a hybrid architecture in which the LLM remains fixed or slow-updated, but an internal memory module
(or a small retrieval policy head) is updated online using environment feedback, with a
provable two-time-scale separation between fast policy iteration and slow memory evolution.
This would preserve the stability benefits of our reflected-MDP analysis while inheriting the
efficiency and scalability motivations emphasised in Titans/MIRAS \cite{behrouz2024titans,behrouz2025miras} and EM-LLM \citep{fountas2024human}.

More broadly, recent work on test-time training layers and nested optimisation provides a
conceptual bridge between architecture and learning dynamics. Test-time training (TTT)
instantiates sequence layers whose hidden state is itself a learnable model updated during
inference \cite{sun2024ttt}. Nested Learning argues that modern deep models can be viewed as
systems of nested optimisation problems with their own internal workflows, offering a
principled lens on continual learning and catastrophic forgetting \cite{behrouz2025nested}. An
important future direction is to recast SRDP within this nested-optimisation view: the outer
loop corresponds to fixed LLM representations and long-term objectives, while the inner loop
implements read--write reflective adaptation as an online optimisation over an internal memory
state. Such a unification could clarify when reflection-based memory updates are sufficient
for continual learning, when parameter adaptation is necessary, and how to combine both
without destabilising long-horizon control.

Another dimension worth exploring is principled memory consolidation. Recent work on KV-state
consolidation in Transformer architectures (Bottlenecked Transformers) demonstrates that selectively rewriting and
compressing internal memory can improve predictive efficiency and long-horizon reasoning
without expanding context length \cite{oomerjee2025bottlenecked}. From the perspective of SRDP, such
mechanisms provide an architectural realisation of the write operation, transforming episodic
experience into a compact latent state that supports future reflection. This view is further
supported by our work on Agent K, which shows that structured cycles of interaction,
reflection, abstraction, and consolidation enable effective experiential learning without
parameter fine-tuning \cite{Grosnit2024KolbBasedEL}. Together, these results point toward an architecture in which episodic memory is consolidated on a slow time scale
and queried on a fast time scale, aligning architectural dynamics with the SRDP formalism.

Finally, these connections open new theoretical questions. MIRAS links sequence modelling to
online optimisation and retention mechanisms \cite{behrouz2025miras}; integrating SRDP into this
framework may yield sharper characterisations of memory coverage, approximation error, and
the conditions under which reflective memory updates guarantee monotone improvement and
asymptotic optimality. Establishing such results for neural internal memories (rather than
external case stores) would move SRDP toward a unified
neural state machine architecture in which episodic experience, reflection, and control are implemented end-to-end together.

\section{Acknowledgements}
We are grateful to Xue Yan for valuable discussions on reinforcement learning and kernel methods, Siyuan Guo for insights into bandit algorithms and the discovery–exploration–exploitation trade-off, and Huichi Zhou and Yihang Chen and Yuxiang Chen for assistance with implementation and proof-read. We also thank collaborators who reviewed earlier versions of this manuscript and provided helpful feedback.

{\small\bibliographystyle{plainnat}\bibliography{refs}}

@article{berry1997bandit,
  title={Bandit problems with infinitely many arms},
  author={Berry, Donald A and Chen, Robert W and Zame, Alan and Heath, David C and Shepp, Larry A},
  journal={The Annals of Statistics},
  volume={25},
  number={5},
  pages={2103--2116},
  year={1997},
  publisher={Institute of Mathematical Statistics}
}

@article{oomerjee2025bottlenecked,
  title={Bottlenecked Transformers: Periodic KV Cache Abstraction for Generalised Reasoning},
  author={Oomerjee, Adnan and Fountas, Zafeirios and Yu, Zhongwei and Bou-Ammar, Haitham and Wang, Jun},
  journal={arXiv preprint arXiv:2505.16950},
  year={2025}
}

@book{tulving1983elements,
  title     = {Elements of Episodic Memory},
  author    = {Tulving, Endel},
  year      = {1983},
  publisher = {Oxford University Press},
  address   = {Oxford, UK}
}

@article{yang2025agentic,
  title={Agentic Web: Weaving the Next Web with AI Agents},
  author={Yang, Yingxuan and Ma, Mulei and Huang, Yuxuan and Chai, Huacan and Gong, Chenyu and Geng, Haoran and Zhou, Yuanjian and Wen, Ying and Fang, Meng and Chen, Muhao and Gu, Shangding and Jin, Ming and Spanos, Costas and Yang, Yang and Abbeel, Pieter and Song, Dawn and Zhang, Weinan and Wang, Jun},
  journal={arXiv preprint arXiv:2507.21206},
  year={2025}
}

@article{ouyang2022training,
  title={Training language models to follow instructions with human feedback},
  author={Ouyang, Long and Wu, Jeffrey and Jiang, Xu and Almeida, Diogo and Wainwright, Carroll and Mishkin, Pamela and Zhang, Chong and Agarwal, Sandhini and Slama, Katarina and Ray, Alex and others},
  journal={Advances in neural information processing systems},
  volume={35},
  pages={27730--27744},
  year={2022}
}

@article{turing1996intelligent,
  title={Intelligent machinery, a heretical theory},
  author={Turing, AM},
  journal={Philosophia Mathematica},
  volume={4},
  number={3},
  year={1996}
}

@article{silver2025welcome,
  title={Welcome to the era of experience},
  author={Silver, David and Sutton, Richard S},
  journal={Google AI},
  volume={1},
  year={2025}
}

@inproceedings{lightman2023let,
  title={Let's verify step by step},
  author={Lightman, Hunter and Kosaraju, Vineet and Burda, Yuri and Edwards, Harrison and Baker, Bowen and Lee, Teddy and Leike, Jan and Schulman, John and Sutskever, Ilya and Cobbe, Karl},
  booktitle={The Twelfth International Conference on Learning Representations},
  year={2023}
}

@article{wang2024openr,
  title={Open{R}: An open source framework for advanced reasoning with large language models},
  author={Wang, Jun and Fang, Meng and Wan, Ziyu and Wen, Muning and Zhu, Jiachen and Liu, Anjie and Gong, Ziqin and Song, Yan and Chen, Lei and Ni, Lionel M and others},
  journal={arXiv preprint arXiv:2410.09671},
  year={2024}
}

@article{wang2025tutorial,
  title={A tutorial on {LLM} reasoning: Relevant methods behind chatgpt o1},
  author={Wang, Jun},
  journal={arXiv preprint arXiv:2502.10867},
  year={2025}
}

@inproceedings{wan2024alphazero,
  title={AlphaZero-Like Tree-Search can Guide Large Language Model Decoding and Training},
  author={Wan, Ziyu and Feng, Xidong and Wen, Muning and Mcaleer, Stephen Marcus and Wen, Ying and Zhang, Weinan and Wang, Jun},
  booktitle={International Conference on Machine Learning},
  pages={49890--49920},
  year={2024},
  organization={PMLR}
}

@inproceedings{zeng2024token,
  title={Token-level Direct Preference Optimization},
  author={Zeng, Yongcheng and Liu, Guoqing and Ma, Weiyu and Yang, Ning and Zhang, Haifeng and Wang, Jun},
  booktitle={International Conference on Machine Learning},
  pages={58348--58365},
  year={2024},
  organization={PMLR}
}

@article{mnih2015human,
  title={Human-level control through deep reinforcement learning},
  author={Mnih, Volodymyr and Kavukcuoglu, Koray and Silver, David and Rusu, Andrei A and Veness, Joel and Bellemare, Marc G and Graves, Alex and Riedmiller, Martin and Fidjeland, Andreas K and Ostrovski, Georg and others},
  journal={nature},
  volume={518},
  number={7540},
  pages={529--533},
  year={2015},
  publisher={Nature Publishing Group}
}

@article{guo2025deepseek,
  title={Deepseek-{R}1: Incentivizing reasoning capability in {LLM}s via reinforcement learning},
  author={Guo, Daya and Yang, Dejian and Zhang, Haowei and Song, Junxiao and Zhang, Ruoyu and Xu, Runxin and Zhu, Qihao and Ma, Shirong and Wang, Peiyi and Bi, Xiao and Zhang, Xiaokang and Yu, Xingkai and Wu, Yu and Wu, Z. F. and Gou, Zhibin and Shao, Zhihong and others},
  journal={arXiv preprint arXiv:2501.12948},
  year={2025}
}

@book{goodfellow2016deep,
  title={Deep learning},
  author={Goodfellow, Ian and Bengio, Yoshua and Courville, Aaron and Bengio, Yoshua},
  year={2016},
  publisher={MIT press Cambridge}
}

@article{rumelhart1986learning,
  title={Learning representations by back-propagating errors},
  author={Rumelhart, David E and Hinton, Geoffrey E and Williams, Ronald J},
  journal={nature},
  volume={323},
  number={6088},
  pages={533--536},
  year={1986},
  publisher={Nature Publishing Group UK London}
}

@article{christianos2023pangu,
  title={Pangu-agent: A fine-tunable generalist agent with structured reasoning},
  author={Christianos, Filippos and Papoudakis, Georgios and Zimmer, Matthieu and Coste, Thomas and Wu, Zhihao and Chen, Jingxuan and Khandelwal, Khyati and Doran, James and Feng, Xidong and Liu, Jiacheng and Zheng Xiong and Yicheng Luo and Jianye Hao and Kun Shao and Haitham Bou-Ammar and Jun Wang },
  journal={arXiv preprint arXiv:2312.14878},
  year={2023}
}

@inproceedings{yao2023tree,
  title     = {Tree of Thoughts: Deliberate Problem Solving with Large Language Models},
  author    = {Yao, Shunyu and Yu, Dian and Zhao, Jeffrey and Shafran, Izhak and Griffiths, Thomas L. and Cao, Yuan and Narasimhan, Karthik},
  booktitle = {Proceedings of the 37th International Conference on Neural Information Processing Systems},
  year      = {2023}
}

@inproceedings{besta2024graph,
  title        = {Graph of Thoughts: Solving Elaborate Problems with Large Language Models},
  author       = {Besta, Maciej and Blach, Nils and Kubicek, Ales and Gerstenberger, Robert and Gianinazzi, Lukas and Gajda, Joanna and Lehmann, Tomasz and Podstawski, Micha{\l} and Niewiadomski, Hubert and Nyczyk, Piotr and Hoefler, Torsten},
  booktitle    = {Proceedings of the Thirty-Eighth {AAAI} Conference on Artificial Intelligence},
  year         = {2024},
}

@inproceedings{guo2025optimizing,
  title={Optimizing case-based reasoning system for functional test script generation with large language models},
  author={Guo, Siyuan and Liu, Huiwu and Chen, Xiaolong and Xie, Yuming and Zhang, Liang and Han, Tao and Chen, Hechang and Chang, Yi and Wang, Jun},
  booktitle={Proceedings of the 31st ACM SIGKDD Conference on Knowledge Discovery and Data Mining V. 2},
  pages={4487--4498},
  year={2025}
}

@article{wang2008unified,
  title={Unified relevance models for rating prediction in collaborative filtering},
  author={Wang, Jun and De Vries, Arjen P and Reinders, Marcel JT},
  journal={ACM Transactions on Information Systems (TOIS)},
  volume={26},
  number={3},
  pages={1--42},
  year={2008},
  publisher={ACM New York, NY, USA}
}

@article{BorkarMeyn2000,
  title={The {ODE} method for convergence of stochastic approximation and reinforcement learning},
  author={Borkar, Vivek S. and Meyn, Sean P.},
  journal={SIAM Journal on Control and Optimization},
  volume={38},
  number={2},
  pages={447--469},
  year={2000}
}

@inproceedings{guo2024ds,
  title={{DS}-{A}gent: Automated Data Science by Empowering Large Language Models with Case-Based Reasoning},
  author={Guo, Siyuan and Deng, Cheng and Wen, Ying and Chen, Hechang and Chang, Yi and Wang, Jun},
  booktitle={International Conference on Machine Learning},
  pages={16813--16848},
  year={2024},
  organization={PMLR}
}

@article{blundell2016model,
  title={Model-free episodic control},
  author={Blundell, Charles and Uria, Benigno and Pritzel, Alexander and Li, Yazhe and Ruderman, Avraham and Leibo, Joel Z and Rae, Jack and Wierstra, Daan and Hassabis, Demis},
  journal={arXiv preprint arXiv:1606.04460},
  year={2016}
}

@inproceedings{pritzel2017neural,
  title={Neural episodic control},
  author={Pritzel, Alexander and Uria, Benigno and Srinivasan, Sriram and Badia, Adria Puigdomenech and Vinyals, Oriol and Hassabis, Demis and Wierstra, Daan and Blundell, Charles},
  booktitle={International conference on machine learning},
  pages={2827--2836},
  year={2017},
  organization={PMLR}
}

@article{ramani2019short,
  title={A short survey on memory based reinforcement learning},
  author={Ramani, Dhruv},
  journal={arXiv preprint arXiv:1904.06736},
  year={2019}
}

@article{hollmann2025accurate,
  title={Accurate predictions on small data with a tabular foundation model},
  author={Hollmann, Noah and M{\"u}ller, Samuel and Purucker, Lennart and Krishnakumar, Arjun and K{\"o}rfer, Max and Hoo, Shi Bin and Schirrmeister, Robin Tibor and Hutter, Frank},
  journal={Nature},
  volume={637},
  number={8045},
  pages={319--326},
  year={2025},
  publisher={Nature Publishing Group UK London}
}

@inproceedings{driess2023palm,
author = {Driess, Danny and Xia, Fei and Sajjadi, Mehdi S. M. and Lynch, Corey and Chowdhery, Aakanksha and Ichter, Brian and Wahid, Ayzaan and Tompson, Jonathan and Vuong, Quan and Yu, Tianhe and Huang, Wenlong and Chebotar, Yevgen and Sermanet, Pierre and Duckworth, Daniel and Levine, Sergey and Vanhoucke, Vincent and Hausman, Karol and Toussaint, Marc and Greff, Klaus and Zeng, Andy and Mordatch, Igor and Florence, Pete},
title = {PaLM-{E}: an embodied multimodal language model},
year = {2023},
booktitle = {Proceedings of the 40th International Conference on Machine Learning},
}

@article{hochreiter1997long,
  title   = {Long Short-Term Memory},
  author  = {Hochreiter, Sepp and Schmidhuber, J{\"u}rgen},
  journal = {Neural Computation},
  volume  = {9},
  number  = {8},
  pages   = {1735--1780},
  year    = {1997}
}

@article{schaul2015prioritized,
  title   = {Prioritized Experience Replay},
  author  = {Schaul, Tom and Quan, John and Antonoglou, Ioannis and others},
  journal = {arXiv preprint arXiv:1511.05952},
  year    = {2015}
}

@article{haarnoja2018soft,
  title   = {Soft {A}ctor-{C}ritic: off-Policy Maximum Entropy Deep Reinforcement Learning with a Stochastic Actor},
  author  = {Haarnoja, Tuomas and Zhou, Aurick and Hartikainen, Kristian and others},
  journal = {arXiv preprint arXiv:1801.01290},
  year    = {2018}
}

@article{schulman2015trust,
  title   = {Trust Region Policy Optimization},
  author  = {Schulman, John and Levine, Sergey and Abbeel, Pieter and others},
  journal = {Proceedings of the 32nd International Conference on Machine Learning},
  year    = {2015}
}

@article{abdolmaleki2018maximum,
  title   = {Maximum a Posteriori Policy Optimisation},
  author  = {Abdolmaleki, Abbas and Springenberg, Jost Tobias and Degrave, Jonas and others},
  journal = {arXiv preprint arXiv:1806.06920},
  year    = {2018}
}

@article{neu2017unified,
  title   = {A Unified View of Entropy-Regularized Markov Decision Processes},
  author  = {Neu, Gergely and Jonsson, Anders and G{\'o}mez, Vicen{\c{c}}},
  journal = {arXiv preprint arXiv:1705.07798},
  year    = {2017}
}

@book{borkar2009stochastic,
  title     = {Stochastic Approximation: A Dynamical Systems Viewpoint},
  author    = {Borkar, Vivek S.},
  publisher = {Springer},
  year      = {2009}
}

@inproceedings{hadamard2024,
  title   = {Stable Hadamard Memory: Revitalizing Memory-Augmented Agents for Reinforcement Learning},
  author  = {Hung Le and Dung Nguyen and Kien Do and Sunil Gupta and Svetha Venkatesh},
  booktitle = {Proc.\ of ICLR 2025},
  year    = {2025},
  archivePrefix = {arXiv},
  eprint  = {2410.10132},
}

@article{hami2025,
  title   = {A scalable reinforcement learning framework inspired by hippocampal memory mechanisms for efficient contextual and sequential decision making},
  author  = {H. Poursiami and A. Moshruba and K. W. Cooper and et al.},
  journal = {Scientific Reports},
  volume  = {15},
  pages   = {25221},
  year    = {2025},
  doi     = {10.1038/s41598-025-10586-x},
}

@book{shalev2014understanding,
  title={Understanding machine learning: From theory to algorithms},
  author={Shalev-Shwartz, Shai and Ben-David, Shai},
  year={2014},
  publisher={Cambridge university press}
}

@article{zhou2025memento,
  title={Memento: Fine-tuning {LLM} agents without fine-tuning {LLM}s},
  author={Zhou, Huichi and Chen, Yihang and Guo, Siyuan and Yan, Xue and Lee, Kin Hei and Wang, Zihan and Lee, Ka Yiu and Zhang, Guchun and Shao, Kun and Yang, Linyi and Wang, Jun},
  journal={Preprint},
  year={2025}
}

@article{brown2020language,
  title={Language models are few-shot learners},
  author={Brown, Tom and Mann, Benjamin and Ryder, Nick and Subbiah, Melanie and Kaplan, Jared D and Dhariwal, Prafulla and Neelakantan, Arvind and Shyam, Pranav and Sastry, Girish and Askell, Amanda and others},
  journal={Advances in neural information processing systems},
  volume={33},
  pages={1877--1901},
  year={2020}
}

@article{valiant1984theory,
  title={A theory of the learnable},
  author={Valiant, Leslie G},
  journal={Communications of the ACM},
  volume={27},
  number={11},
  pages={1134--1142},
  year={1984},
  publisher={ACM New York, NY, USA}
}

@article{McClelland1995CLS,
  title   = {Why There Are Complementary Learning Systems in the Hippocampus and Neocortex: Insights from the Successes and Failures of Connectionist Models of Learning and Memory},
  author  = {McClelland, James L. and McNaughton, Bruce L. and O'Reilly, Randall C.},
  journal = {Psychological Review},
  volume  = {102},
  number  = {3},
  pages   = {419--457},
  year    = {1995},
  doi     = {10.1037/0033-295X.102.3.419}
}

@inproceedings{Wei2022CoT,
  title     = {Chain-of-Thought Prompting Elicits Reasoning in Large Language Models},
  author    = {Wei, Jason and Wang, Xuezhi and Schuurmans, Dale and Bosma, Maarten and Ichter, Brian and Xia, Fei and Chi, Ed H. and Le, Quoc V. and Zhou, Denny},
  booktitle = {Advances in Neural Information Processing Systems (NeurIPS)},
  year      = {2022},
 }

@article{Yao2022ReAct,
  title   = {ReAct: Synergizing Reasoning and Acting in Language Models},
  author  = {Yao, Shunyu and Zhao, Jeffrey and Yu, Dian and Du, Nan and Shafran, Izhak and Narasimhan, Karthik and Cao, Yuan},
  journal = {arXiv preprint arXiv:2210.03629},
  year    = {2022},
}

@inproceedings{Lewis2020RAG,
  title     = {Retrieval-Augmented Generation for Knowledge-Intensive {NLP} Tasks},
  author    = {Lewis, Patrick and Perez, Ethan and Piktus, Aleksandra and Petroni, Fabio and Karpukhin, Vladimir and Goyal, Naman and K{\"u}ttler, Heinrich and Lewis, Mike and Yih, Wen-tau and Rockt{\"a}schel, Tim and others},
  booktitle = {Advances in Neural Information Processing Systems (NeurIPS)},
 year      = {2020},
 }

@inproceedings{Borgeaud2022RETRO,
  title     = {Improving Language Models by Retrieving from Trillions of Tokens},
  author    = {Borgeaud, Sebastian and Mensch, Arthur and Hoffmann, Jordan and Cai, Trevor and Rutherford, Eliza and Millican, Katie and Van Den Driessche, George and Lespiau, Jean-Baptiste and Damoc, Bogdan and Clark, Aidan and de Las Casas, Diego and Guy, Aur{\'e}lia and Menick, Jacob and Ring, Roman and Hennigan, Tom and Huang, Saffron and Maggiore, Loren and Jones, Chris and Cassirer, Albin and Brock, Andy and Paganini, Michela and Irving, Geoffrey and Vinyals, Oriol and Osindero, Simon and Simonyan, Karen and Rae, Jack and Elsen, Erich and Sifre, Laurent},
  booktitle = {Proceedings of the 39th International Conference on Machine Learning (ICML)},
  year={2022}
}

@article{GershmanDaw2017EpisodicRL,
  title   = {Reinforcement Learning and Episodic Memory in Humans and Animals: An Integrative Framework},
  author  = {Gershman, Samuel J. and Daw, Nathaniel D.},
  journal = {Annual Review of Psychology},
  volume  = {68},
  pages   = {101--128},
  year    = {2017},
  doi     = {10.1146/annurev-psych-122414-033625}
}

@book{BertsekasTsitsiklis1996,
  title     = {Neuro-Dynamic Programming},
  author    = {Bertsekas, Dimitri P. and Tsitsiklis, John N.},
  year      = {1996},
  publisher = {Athena Scientific}
}

@book{Puterman1994,
  title     = {Markov Decision Processes: Discrete Stochastic Dynamic Programming},
  author    = {Puterman, Martin L.},
  year      = {1994},
  publisher = {Wiley}
}

@book{SuttonBarto2018,
  title     = {Reinforcement Learning: An Introduction},
  author    = {Sutton, Richard S. and Barto, Andrew G.},
  edition   = {2},
  year      = {2018},
  publisher = {MIT Press}
}

@book{BoydVandenberghe2004,
  title     = {Convex Optimization},
  author    = {Boyd, Stephen and Vandenberghe, Lieven},
  year      = {2004},
  publisher = {Cambridge University Press}
}

@article{AamodtPlaza1994,
  title   = {Case-Based Reasoning: foundational Issues, Methodological Variations, and System Approaches},
  author  = {Aamodt, Agnar and Plaza, Enric},
  journal = {AI Communications},
  year    = {1994},
  volume  = {7},
  number  = {1},
  pages   = {39--59}
}

@book{Ashley1990,
  title     = {Modeling Legal Argument: Reasoning with Cases and Hypotheticals},
  author    = {Ashley, Kevin D.},
  year      = {1990},
  publisher = {MIT Press}
}

@article{yu2025self,
  title={Self-Verifying Reflection Helps Transformers with CoT Reasoning},
  author={Yu, Zhongwei and Xia, Wannian and Yan, Xue and Xu, Bo and Zhang, Haifeng and Du, Yali and Wang, Jun},
  journal={arXiv preprint arXiv:2510.12157},
  year={2025}
}

@article{Grosnit2024KolbBasedEL,
  title={Kolb-Based Experiential Learning for Generalist Agents with Human-Level Kaggle Data Science Performance},
  author={Antoine Grosnit and Alexandre Max Maraval and N RefinathS and Zichao Zhao and James Doran and Giuseppe Paolo and Albert Thomas and Jonas Gonzalez and Abhineet Kumar and Khyati Khandelwal and Abdelhakim Benechehab and Hamza Cherkaoui and Youssef Attia El-Hili and Kun Shao and Jianye Hao and Jun Yao and Bal{\'a}zs K{\'e}gl and Haitham Bou-Ammar and Jun Wang},
  year={2025},
  journal={arXiv preprint}
}

@article{kaelbling1998planning,
author = {Kaelbling, Leslie Pack and Littman, Michael L. and Cassandra, Anthony R.},
title = {Planning and acting in partially observable stochastic domains},
year = {1998},
journal = {Artif. Intell.},
}

@inproceedings{hausknecht2015deep,
  title={Deep Recurrent {Q}-Learning for Partially Observable MDPs.},
  author={Hausknecht, Matthew J and Stone, Peter},
  booktitle={AAAI fall symposia},
  volume={45},
  pages={141},
  year={2015}
}

@article{behrouz2024titans,
  title        = {Titans: Learning to Memorize at Test Time},
  author       = {Behrouz, Ali and Zhong, Peilin and Mirrokni, Vahab},
  journal      = {arXiv preprint arXiv:2501.00663},
  year         = {2024}
}

@article{behrouz2025miras,
  title        = {It's All Connected: A Journey Through Test-Time Memorization, Attentional Bias, Retention, and Online Optimization},
  author       = {Behrouz, Ali and Razaviyayn, Meisam and Zhong, Peilin and Mirrokni, Vahab},
  journal      = {arXiv preprint arXiv:2504.13173},
  year         = {2025}
}

@article{sun2024ttt,
  title        = {Learning to (Learn at Test Time): {RNN}s with Expressive Hidden States},
  author       = {Sun, Yu and Li, Xinhao and Dalal, Karan and Xu, Jiarui and Vikram, Arjun and Zhang, Genghan and Dubois, Yann and Chen, Xinlei and Wang, Xiaolong and Koyejo, Sanmi and Hashimoto, Tatsunori and Guestrin, Carlos},
  journal      = {arXiv preprint arXiv:2407.04620},
  year         = {2024}
}

@inproceedings{behrouz2025nested,
  title        = {Nested Learning: The Illusion of Deep Learning Architectures},
  author       = {Behrouz, Ali and Razaviyayn, Meisam and Zhong, Peilin and Mirrokni, Vahab},
  booktitle    = {Advances in Neural Information Processing Systems},
  year         = {2025}
}

@article{Ziebart2010,
  title   = {Modeling Purposeful Adaptive Behavior with the Principle of Maximum Causal Entropy},
  author  = {Ziebart, Brian D.},
  journal = {PhD thesis, Carnegie Mellon University},
  year    = {2010}
}

@article{fountas2024human,
  title   = {Human-like episodic memory for infinite context LLMs},
  author  = {Fountas, Zafeirios and Benfeghoul, Martin A. and Oomerjee, Adnan and Christopoulou, Fenia and Lampouras, Gerasimos and Bou-Ammar, Haitham and Wang, Jun},
  journal = {arXiv preprint arXiv:2407.09450},
  year    = {2024}
}

@inproceedings{Zhong2024,
  title     = {MemoryBank: Enhancing Large Language Models with Long-Term Memory},
  author    = {Zhong, Wanjun and Guo, Lianghong and Gao, Qiqi and Ye, He and Wang, Yanlin},
  booktitle = {Proceedings of the AAAI Conference on Artificial Intelligence},
  year      = {2024},
  volume    = {38},
  pages     = {19724--19731}
}

@article{Chhikara2025,
  title   = {Mem0: Building Production-Ready {AI} Agents with Scalable Long-Term Memory},
  author  = {Chhikara, Prateek and Khant, Dev and Aryan, Saket and Singh, Taranjeet and Yadav, Deshraj},
  journal = {arXiv preprint arXiv:2504.19413},
  year    = {2025}
}

@article{Parzen1962,
  title   = {On Estimation of a Probability Density Function and Mode},
  author  = {Parzen, Emanuel},
  journal = {The Annals of Mathematical Statistics},
  year    = {1962},
  volume  = {33},
  number  = {3},
  pages   = {1065--1076}
}

\appendix
\section{Proof of Lemma~\ref{lemma-objective}}
\label{appendix:proofs}

\begin{proof}
The Lagrangian for \eqref{eq:parzen-kl-objective} with the constraint $\sum_c \nu(c)=1$ is
\[
L(\nu,\lambda)=\sum_c \nu(c)Q(x,c)\;-\;\alpha\sum_c \nu(c)\log\frac{\nu(c)}{\mu_0(c\mid x)}
\;-\;\lambda\Big(\sum_c \nu(c)-1\Big).
\]

Taking the partial derivative w.r.t.\ $\nu(c)$ and setting it to zero, we use
\[
\frac{\partial}{\partial \nu}\Big[\nu \log(\nu/\mu_0)\Big]
= \log(\nu/\mu_0)+1,
\]
to obtain
\[
\frac{\partial L}{\partial \nu(c)}
= Q(x,c) - \alpha\Big(\log\frac{\nu(c)}{\mu_0(c\mid x)}+1\Big) - \lambda
=0.
\]
Equivalently,
\[
\log\frac{\nu(c)}{\mu_0(c\mid x)}
=\frac{Q(x,c)-\lambda}{\alpha}-1.
\]
Therefore,
\[
\nu(c)=\mu_0(c\mid x)\exp\!\Big(\tfrac{1}{\alpha}Q(x,c)\Big)\exp\!\Big(-1-\tfrac{\lambda}{\alpha}\Big).
\]
Imposing $\sum_c \nu(c)=1$ gives the normalising constant
\[
\exp\!\Big(-1-\tfrac{\lambda}{\alpha}\Big)
=
\left(\sum_c \mu_0(c\mid x)\exp\!\Big(\tfrac{1}{\alpha}Q(x,c)\Big)\right)^{-1}.
\]
Substituting back yields \eqref{eq:parzen-kl-solution}.
\end{proof}

\section{Proof of Theorem \ref{thm:parzen-spi-convergence}}
Unlike standard soft reinforcement learning~\citep{haarnoja2018soft}, our setting involves state-dependent base measures that vary with memory content, requiring a careful analysis of the induced operators. In particular, extending KL-regularised evaluation requires establishing that the associated log-sum-exp operations preserve contraction for both the value function and the prior distributed resulted from memory updates. Lemma~\ref{lem:lse-bi-lip} extends the log-sum-exp Lipschitz bound~\cite{BoydVandenberghe2004} to this setting with Parzen base measures and memory updates.
\subsection{Bi-Lipschitz Properties}

\begin{lemma}[Bi-Lipschitz property of log-sum-exp with base measure]
\label{lem:lse-bi-lip}
Let $q,q'\in\mathbb{R}^n$, $b,b'\in\Delta_n$, and $\alpha>0$. Define
\[
f(b,q)\;:=\;\alpha\log\sum_{i=1}^n b_i \exp(q_i/\alpha).
\]
Then:
\begin{enumerate}
\item \textbf{(Lipschitz in $q$)} For any fixed $b\in\Delta_n$,
\[
|f(b,q)-f(b,q')|\le \|q-q'\|_\infty.
\]
\item \textbf{(Lipschitz in $b$)} If $q_i\in[Q_{\min},Q_{\max}]$ for all $i$ and
$\Delta_Q=Q_{\max}-Q_{\min}$, then for any fixed $q$,
\[
|f(b,q)-f(b',q)|\le \alpha e^{\Delta_Q/\alpha}\,\|b-b'\|_1
= 2\alpha e^{\Delta_Q/\alpha}\,\mathrm{TV}(b,b').
\]
\item \textbf{(Joint bound)} Under the same boundedness condition,
\[
|f(b,q)-f(b',q')|
\le
\|q-q'\|_\infty + \alpha e^{\Delta_Q/\alpha}\,\|b-b'\|_1.
\]
\end{enumerate}
\end{lemma}

\begin{proof}
\textbf{(1) Lipschitz in $q$ (base-measure log-sum-exp).}
Fix any $b\in\Delta_n$ and define the single-argument function
\[
f(q)\;:=\;f(b,q)=\alpha \log \sum_{i=1}^n b_i \exp\!\left(\frac{q_i}{\alpha}\right).
\]
The function $f$ is differentiable with gradient
\[
\nabla f(q)
=
\left[
\frac{b_i \exp(q_i/\alpha)}{\sum_{j=1}^n b_j \exp(q_j/\alpha)}
\right]_{i=1}^n,
\]
which is a probability vector (all components are nonnegative and sum to one), hence
\[
\|\nabla f(q)\|_1 = 1 \qquad \text{for all } q\in\mathbb{R}^n.
\]
By the mean value theorem for vector-valued functions (e.g., \cite{BoydVandenberghe2004}),
there exists $\xi$ on the line segment between $q$ and $q'$ such that
\[
f(q)-f(q')=\nabla f(\xi)^\top (q-q').
\]
Therefore, by Hölder's inequality,
\[
|f(b,q)-f(b,q')|
=
|f(q)-f(q')|
\le
\|\nabla f(\xi)\|_1\,\|q-q'\|_\infty
=
\|q-q'\|_\infty.
\]

\textbf{(2) Lipschitz in $b$.}
Fix any $q$ and write $f(b,q)=\alpha\log(b^\top w)$ with $w_i=e^{q_i/\alpha}$.
Then
\[
\nabla_b f(b,q)=\alpha\,\frac{w}{b^\top w}.
\]
If $q_i\in[Q_{\min},Q_{\max}]$ for all $i$, then
\[
\min_i w_i=e^{Q_{\min}/\alpha},\qquad \max_i w_i=e^{Q_{\max}/\alpha},
\qquad b^\top w \ge \min_i w_i=e^{Q_{\min}/\alpha}.
\]
Hence
\[
\|\nabla_b f(b,q)\|_\infty
\le
\alpha\,\frac{\max_i w_i}{\min_i w_i}
=
\alpha e^{(Q_{\max}-Q_{\min})/\alpha}
=
\alpha e^{\Delta_Q/\alpha}.
\]
Applying the mean value theorem in $b$ and $\ell_1$--$\ell_\infty$ duality gives
\[
|f(b,q)-f(b',q)|
\le
\|\nabla_b f(\xi,q)\|_\infty \,\|b-b'\|_1
\le
\alpha e^{\Delta_Q/\alpha}\,\|b-b'\|_1
=
2\alpha e^{\Delta_Q/\alpha}\,\mathrm{TV}(b,b').
\]

\textbf{(3) Joint bound.}
Add and subtract $f(b,q')$ and apply (1)--(2):
\[
|f(b,q)-f(b',q')|
\le
|f(b,q)-f(b,q')|
+
|f(b,q')-f(b',q')|
\le
\|q-q'\|_\infty + \alpha e^{\Delta_Q/\alpha}\,\|b-b'\|_1.
\]
\end{proof}

Next, we use the above to establish the contraction for the KL-regularised evaluation. 
\subsection{Contraction of the KL-Regularised Evaluation Operator}
\label{subsec:kl-eval-contraction}

We prove Theorem~\ref{thm:kl-eval-contraction} using a single unified Lipschitz lemma
for log-sum-exp with a base measure (Lemma~\ref{lem:lse-bi-lip}), which simultaneously
provides (i) Lipschitz continuity in the value argument $Q$ and (ii) stability with
respect to perturbations of the base measure induced by memory.

\begin{theorem}[Contraction of KL-Regularised Evaluation]
\label{thm:kl-eval-contraction}
Fix a memory state $M$ and a base measure $\mu_0^M(\cdot\mid x)$ for each context $x$.
Define the KL-regularised evaluation operator $\mathcal{T}^{\mu,M}_{\mathrm{KL}}$
acting on $Q:\mathcal{X}\times\mathcal{C}\to\mathbb{R}$ by
\begin{equation}
\label{eq:kl-eval-operator-withM}
(\mathcal{T}^{\mu,M}_{\mathrm{KL}}Q)(x,c)
\;:=\;
r(x,c) + \gamma\,\alpha\log\!\Big(\sum_{c'\in\mathcal{C}}
\mu_0^M(c'\mid x')\exp(Q(x',c')/\alpha)\Big),
\end{equation}
where $x'$ denotes the next context/state (possibly random) following $(x,c)$,
and $\gamma\in(0,1)$ is the discount factor.
Then $\mathcal{T}^{\mu,M}_{\mathrm{KL}}$ is a $\gamma$-contraction in the
$\|\cdot\|_\infty$ norm:
\[
\|\mathcal{T}^{\mu,M}_{\mathrm{KL}}Q - \mathcal{T}^{\mu,M}_{\mathrm{KL}}Q'\|_\infty
\;\le\;
\gamma \|Q-Q'\|_\infty.
\]
Consequently, $\mathcal{T}^{\mu,M}_{\mathrm{KL}}$ admits a unique fixed point $Q_M^\star$.
\end{theorem}

\begin{proof}
Fix $M$ and let $Q,Q'$ be arbitrary bounded functions.
For any $(x,c)$, subtract \eqref{eq:kl-eval-operator-withM} evaluated at $Q$ and $Q'$:
\begin{align*}
&\big|(\mathcal{T}^{\mu,M}_{\mathrm{KL}}Q)(x,c)
-(\mathcal{T}^{\mu,M}_{\mathrm{KL}}Q')(x,c)\big| \\
&\qquad=
\gamma \left|
\alpha\log\!\Big(\sum_{c'} \mu_0^M(c'\mid x')e^{Q(x',c')/\alpha}\Big)
-
\alpha\log\!\Big(\sum_{c'} \mu_0^M(c'\mid x')e^{Q'(x',c')/\alpha}\Big)
\right|.
\end{align*}
Define, for the (fixed) probability vector $b=\mu_0^M(\cdot\mid x')$,
\[
f(b,q) := \alpha\log\sum_{i} b_i e^{q_i/\alpha},
\qquad
q := Q(x',\cdot),\quad q' := Q'(x',\cdot).
\]
By Lemma~\ref{lem:lse-bi-lip}(1) (Lipschitz in $q$ for fixed $b$), we have
\[
\left| f(b,q)-f(b,q') \right| \le \|q-q'\|_\infty
=
\|Q(x',\cdot)-Q'(x',\cdot)\|_\infty
\le
\|Q-Q'\|_\infty.
\]
Therefore, for all $(x,c)$,
\[
\big|(\mathcal{T}^{\mu,M}_{\mathrm{KL}}Q)(x,c)
-(\mathcal{T}^{\mu,M}_{\mathrm{KL}}Q')(x,c)\big|
\le
\gamma \|Q-Q'\|_\infty.
\]
Taking the supremum over $(x,c)$ yields
\[
\|\mathcal{T}^{\mu,M}_{\mathrm{KL}}Q - \mathcal{T}^{\mu,M}_{\mathrm{KL}}Q'\|_\infty
\le
\gamma \|Q-Q'\|_\infty,
\]
so $\mathcal{T}^{\mu,M}_{\mathrm{KL}}$ is a $\gamma$-contraction.
Existence and uniqueness of the fixed point follow from Banach's fixed-point theorem.
\end{proof}

\begin{remark}[Stability under memory perturbations]
\label{rem:kl-eval-memory-stability}
The same Lemma~\ref{lem:lse-bi-lip} also provides stability of the evaluation operator
to changes in memory-induced base measures. In particular, under boundedness of $Q$,
Lemma~\ref{lem:lse-bi-lip}(2) implies that for any $M,M'$,
\[
\sup_{x'}\left|
\alpha\log\!\sum_{c'} \mu_0^{M}(c'\mid x')e^{Q(x',c')/\alpha}
-
\alpha\log\!\sum_{c'} \mu_0^{M'}(c'\mid x')e^{Q(x',c')/\alpha}
\right|
\le
2\alpha e^{\Delta_Q/\alpha}\,\|M'-M\|.
\]
This perturbation bound is used later to establish the Lipschitz continuity of the
moving fixed point $Q_M^\star$ and to derive the tracking recursion for two-timescale
read--write learning.
\end{remark}

We next establish monotone improvement results for \emph{Read--Write Learning Iterations}
(Theorem~\ref{thm:parzen-improvement}). 

\subsection{Monotone Improvement of Read-Write Iterations}
Our analysis relies on the
\emph{Gibbs variational principle}, also known as the log-sum-exp conjugate duality
\citep[Section~3.3]{BoydVandenberghe2004}, which connects entropy-regularised optimisation
to exponential-family distributions. In particular, the Gibbs principle states that
\[
\alpha \log \sum_i e^{q_i/\alpha}
=
\max_{\nu \in \Delta_n}
\left\{
\sum_i \nu_i q_i - \alpha H(\nu)
\right\},
\]
where $H(\nu) = -\sum_i \nu_i \log \nu_i$ denotes the Shannon entropy. This identity
characterises the log-sum-exp operator as the optimal value of an entropy-regularised
linear objective.

Our setting extends this principle by incorporating a \emph{base measure} through the
Kullback--Leibler divergence $\mathrm{KL}(\nu \| \mu_0)$, replacing entropy regularisation
with relative entropy. This yields a Parzen-style exponential-family formulation and
corresponds to the Bregman divergence induced by negative entropy
\citep{Ziebart2010}.

\begin{theorem}[Monotone Improvement Under Parzen-KL]
\label{thm:parzen-improvement}
Let $Q^{\mu}$ be the fixed point of $\mathcal{T}^{\mu}_{\mathrm{KL}}$ and define $\mu^{+}$ via \eqref{eq:parzen-kl-solution}. Then:
$$V^{\mu^{+}}(x) \geq V^{\mu}(x) \quad \text{for all } x$$
\end{theorem}

\begin{proof}[Proof:]
By the generalised Gibbs variational principle with base measure $\mu_0$:
\begin{align}
&\alpha\log\sum_{c}\mu_0(c\mid x)e^{Q^\mu(x,c)/\alpha} \\
&= \max_{\nu\in\Delta(M)}\left\{\sum_c \nu(c)Q^\mu(x,c) - \alpha\,\mathrm{KL}(\nu\|\mu_0(\cdot\mid x))\right\}
\end{align}

The maximiser is exactly $\mu^{+}$ from \eqref{eq:parzen-kl-solution}. Therefore:
\begin{align}
V^{\mu^{+}}(x) &= \sum_c \mu^{+}(c\mid x)Q^\mu(x,c) - \alpha\,\mathrm{KL}(\mu^{+}(\cdot\mid x)\|\mu_0(\cdot\mid x)) \\
&= \alpha\log\sum_{c}\mu_0(c\mid x)e^{Q^\mu(x,c)/\alpha} \\
&\geq \sum_c \mu(c\mid x)Q^\mu(x,c) - \alpha\,\mathrm{KL}(\mu(\cdot\mid x)\|\mu_0(\cdot\mid x)) \\
&= V^{\mu}(x)
\end{align}

The inequality follows since $\mu^{+}$ maximises the objective defining $V^{\mu}$.\end{proof}

Finally, we derive the proof of Theorem \ref{thm:parzen-spi-convergence} for the convergence of the \emph{Read-Write Learning}.
\subsection{Convergence of Read-Write Iterations}
\begin{proof}[Proof]
Let $(Q_t,\mu_t)$ be the sequence of iterates. By Theorem \ref{thm:parzen-improvement}, the soft values satisfy $V^{\mu_{t+1}}(x) \geq V^{\mu_t}(x)$ for all $x$.

Since rewards are bounded by $R_{\max}$ and $\gamma < 1$, soft values are uniformly bounded:
$$V^{\mu}(x) \leq \frac{R_{\max}}{1-\gamma} + \alpha\log|M|.$$

The last term uses $\mathrm{KL}(\nu\|\mu_0) \geq -\log|M|$ for any $\nu\in\Delta(M)$. Since $\{V^{\mu_t}\}$ is bounded and monotone increasing, it converges pointwise to some limit $\bar{V}$. By Theorem \ref{thm:kl-eval-contraction}, each evaluation stage converges to the unique $Q^{\mu_t}$. The sequence $\{Q^{\mu_t}\}$ lies in a compact subset of bounded functions. Any limit point $(\bar{Q},\bar{\mu})$ must satisfy:
\begin{itemize}
\item \emph{Bellman consistency}: $\bar{Q} = \mathcal{T}^{\bar{\mu}}_{\mathrm{KL}}\bar{Q}$
\item \emph{KL optimality}: $\bar{\mu}$ maximizes the KL-regularised objective given $\bar{Q}$
\end{itemize}

By continuity of the improvement mapping \eqref{eq:parzen-kl-solution}, these conditions uniquely determine the fixed point, so the entire sequence converges.
\end{proof}

\section{Lipschitz Properties of Memory Updates}
\label{Lipschitz Properties of Memory Updates}
In practice, memory updates typically follow one of these patterns:
\begin{enumerate}
\item \emph{Experience Replay Buffer}:
\begin{align}
M_{t+1} = \begin{cases}
M_t \cup \{(s_t, a_t, r_t)\} & \text{with probability } \rho_t \\
M_t & \text{otherwise}
\end{cases} \label{eq:memory-replay}
\end{align}

\item \emph{Sliding Window Memory}:
\begin{align}
M_{t+1} = (M_t \setminus \{\text{oldest case}\}) \cup \{(s_t, a_t, r_t)\} \quad \text{with probability } \rho_t \label{eq:memory-sliding}
\end{align}

\item \emph{Importance-Weighted Memory}:
\begin{align}
M_{t+1} = M_t \cup \{(s_t, a_t, r_t, w_t)\} \quad \text{where } w_t = f(Q_t, \mu_t) \label{eq:memory-weighted}
\end{align}
\end{enumerate}

Throughout, we make explicit the role of memory as an \emph{operator-valued state}
rather than a raw set. In other words, although memory $M_t$ is implemented as a buffer of cases, the algorithm accesses
memory only through the \emph{induced base measure}
\[
\mu_0^M(\cdot \mid x),
\]
used by the Parzen--KL prior and the KL-regularised evaluation operator.

\begin{definition}[Memory distance induced by base measures]
\label{def:memory-distance}
Let $\mathcal{C}$ be a (countable) universe of case identifiers.  
Each memory buffer $M \subseteq \mathcal{C}$ induces a memory-dependent
base measure
\[
\mu_0^M(\cdot\mid x)\in\Delta_{\mathcal{C}},
\qquad
\text{with }\;
\mu_0^M(c\mid x)=0 \;\;\text{for all } c\notin M .
\]

We define the \emph{memory distance} between two memory buffers $M$ and $M'$
at the operator level as
\begin{equation}
\label{eq:memory-distance-def}
\|M' - M\|
\;:=\;
\sup_{x}
\mathrm{TV}\!\left(
\mu_0^{M'}(\cdot\mid x), \mu_0^{M}(\cdot\mid x)
\right)
\;=\;
\frac12 \sup_x
\big\|
\mu_0^{M'}(\cdot\mid x) - \mu_0^{M}(\cdot\mid x)
\big\|_1 .
\end{equation}

All subsequent norms involving memory are understood in this sense.
\end{definition}

\begin{assumption}[Bounded $Q$-range]
\label{as:Q-range}
For all $x\in\mathcal{X}$ and $c\in\mathcal{C}$,
$Q(x,c)\in[Q_{\min},Q_{\max}]$ and $\Delta_Q:= Q_{\max}-Q_{\min}$.
\end{assumption}

\begin{lemma}[Lipschitz continuity of the Parzen--KL targets in memory]
\label{lem:parzen-targets-lip-memory}
Assume \ref{def:memory-distance}--\ref{as:Q-range}.
Fix any $Q$ and define, for each $x$,
\[
G_M(x;Q)
\;:=\;
\alpha \log \sum_{c'\in \mathcal{C}} \mu_0^M(c'\mid x)\exp\!\big(Q(x,c')/\alpha\big),
\]
and
\[
\Phi_M(c\mid x;Q)
\;:=\;
\frac{\mu_0^M(c\mid x)\exp(Q(x,c)/\alpha)}
{\sum_{c'\in\mathcal{C}}\mu_0^M(c'\mid x)\exp(Q(x,c')/\alpha)}.
\]
Then for any $M,M'$ and all $x$,
\begin{align}
\big|G_M(x;Q)-G_{M'}(x;Q)\big|
&\le
\alpha e^{\Delta_Q/\alpha}\,
\|\mu_0^M(\cdot\mid x)-\mu_0^{M'}(\cdot\mid x)\|_1,
\label{eq:G-lip}\\
\|\Phi_M(\cdot\mid x;Q)-\Phi_{M'}(\cdot\mid x;Q)\|_1
&\le
2 e^{\Delta_Q/\alpha}\,
\|\mu_0^M(\cdot\mid x)-\mu_0^{M'}(\cdot\mid x)\|_1.
\label{eq:Phi-lip}
\end{align}
Consequently,
\[
\sup_x |G_M(x;Q)-G_{M'}(x;Q)|
\le
2\alpha e^{\Delta_Q/\alpha}\,\|M'-M\|,
\]
\[
\sup_x \|\Phi_M(\cdot\mid x;Q)-\Phi_{M'}(\cdot\mid x;Q)\|_1
\le
4 e^{\Delta_Q/\alpha}\,\|M'-M\|.
\]
\end{lemma}

\begin{proof}
Fix $x$ and abbreviate $b=\mu_0^M(\cdot\mid x)$, $b'=\mu_0^{M'}(\cdot\mid x)$, and
$w(c)=\exp(Q(x,c)/\alpha)$.
By Assumption~\ref{as:Q-range}, $w(c)\in[e^{Q_{\min}/\alpha},e^{Q_{\max}/\alpha}]$.

\paragraph{(i) Lipschitzness of $G_M$.}
Write $G_M(x;Q)=\alpha\log(b^\top w)$.
This map is differentiable in $b$ with
$\nabla_b(\alpha\log(b^\top w)) = \alpha\,w/(b^\top w)$.
Since $b^\top w \ge \min_c w(c) = e^{Q_{\min}/\alpha}$,
\[
\|\nabla_b(\alpha\log(b^\top w))\|_\infty
\le
\alpha\,\frac{\max_c w(c)}{\min_c w(c)}
=
\alpha e^{(Q_{\max}-Q_{\min})/\alpha}
=
\alpha e^{\Delta_Q/\alpha}.
\]
By the mean value theorem and H\"older's inequality,
\[
|G_M(x;Q)-G_{M'}(x;Q)|
=
|\nabla_b G_\xi^\top(b-b')|
\le
\|\nabla_b G_\xi\|_\infty \|b-b'\|_1
\le
\alpha e^{\Delta_Q/\alpha}\|b-b'\|_1,
\]
which proves \eqref{eq:G-lip}.

\paragraph{(ii) Lipschitzness of $\Phi_M$.}
Let $w_i=e^{q_i/\alpha}$ and $Z(b)=b^\top w$. Then $\psi(b)_i=b_i w_i/Z(b)$.
Write
\[
\psi(b)-\psi(b') = \frac{b\odot w}{Z(b)}-\frac{b'\odot w}{Z(b')}
=
\underbrace{\frac{(b-b')\odot w}{Z(b)}}_{(A)}
+
\underbrace{(b'\odot w)\Big(\frac{1}{Z(b)}-\frac{1}{Z(b')}\Big)}_{(B)},
\]
where $\odot$ is elementwise product.
For (A), using $Z(b)\ge \min_i w_i=e^{Q_{\min}/\alpha}$,
\[
\|(A)\|_1 \le \frac{\| (b-b')\odot w\|_1}{Z(b)}
\le \frac{\max_i w_i}{\min_i w_i}\,\|b-b'\|_1
= e^{\Delta_Q/\alpha}\|b-b'\|_1.
\]
For (B), note $|Z(b)-Z(b')|\le \|b-b'\|_1 \max_i w_i$ and
$Z(b)Z(b')\ge (\min_i w_i)^2$. Also $\|b'\odot w\|_1=Z(b')\le \max_i w_i$.
Thus
\[
\|(B)\|_1
\le
\|b'\odot w\|_1 \frac{|Z(b)-Z(b')|}{Z(b)Z(b')}
\le
(\max w)\frac{\|b-b'\|_1(\max w)}{(\min w)^2}
=
e^{\Delta_Q/\alpha}\|b-b'\|_1.
\]
Summing the two bounds yields \eqref{eq:Phi-lip}. Finally, applying Definition~\ref{def:memory-distance} gives
$\|b-b'\|_1 \le 2\|M'-M\|$ uniformly over $x$, proving the stated consequences.
\end{proof}

\begin{lemma}[Lipschitz continuity of $F_M$ in memory for updates (14)--(15)]
\label{lem:FM-lip-memory}
Consider the sample-form updates:
\begin{align*}
\delta_t(Q,\mu_0,M)
&:= r_t + \gamma\,\alpha \log\!\Big(\sum_{c'\in M}\mu_0^M(c'\mid x_{t+1})e^{Q(x_{t+1},c')/\alpha}\Big) - Q(x_t,c_t),\\
\Phi_t(c\mid x;Q,\mu_0,M)
&:= \frac{\mu_0^M(c\mid x)e^{Q(x,c)/\alpha}}{\sum_{c'\in M}\mu_0^M(c'\mid x)e^{Q(x,c')/\alpha}}.
\end{align*}
Let $F_M(Z)$ denote the mean-field drift of $(Q,\mu)$ induced by these targets
(i.e., conditional expectations given the current state).
Under~\ref{def:memory-distance}--\ref{as:Q-range}, there exists $L_F<\infty$ such that
for any fixed $Z=(Q,\mu)$,
\[
\|F_{M'}(Z)-F_M(Z)\|
\le
L_F\,\|M'-M\|.
\]
In particular, one may take $L_F$ proportional to $e^{\Delta_Q/\alpha}$.
\end{lemma}

\begin{proof}
Memory enters the TD target only through the log-sum-exp term $G_M(x_{t+1};Q)$,
and enters the policy target only through $\Phi_M(\cdot\mid x_t;Q)$.
By Lemma~\ref{lem:parzen-targets-lip-memory}, both of these are Lipschitz in $M$
under the distance \eqref{def:memory-distance}, uniformly in $x$.
Taking conditional expectations (to pass from sample form to mean-field drift)
preserves Lipschitz constants.
Therefore the combined drift mapping $F_M(Z)$ is Lipschitz in memory, with constant
$L_F$ depending on $\gamma,\alpha$ and the bounded range $\Delta_Q$, and scaling as
$O(e^{\Delta_Q/\alpha})$.
\end{proof}

\section{Proof of Theorem~\ref{thm:twots-parzen-detailed}}
\label{proof:two-time-scale}
We are not ready to prove Theorem~\ref{thm:twots-parzen-detailed} using the ODE method for
two-timescale stochastic approximation
\citep{borkar2009stochastic,BorkarMeyn2000}.

\subsection{Fast Timescale: Fixed-Memory Dynamics}

Fix a memory state $M$. Let $Z=(Q,\mu)$ denote the fast variables, and define the
KL-regularised operator
\[
Z \;\mapsto\; F_M(Z),
\]
which combines KL-regularised policy evaluation and Parzen--KL policy improvement.
The stochastic updates
\eqref{eq:q-update-sample} and \eqref{eq:policy-update-sample}
are stochastic approximations of the limiting ODE
\begin{equation}
\label{eq:fast-ode}
\frac{dZ}{d\tau} = F_M(Z) - Z,
\end{equation}
where $\tau$ denotes fast time.

\begin{assumption}[Uniform Contraction]
\label{as:uniform-contraction}
There exists $\kappa\in(0,1)$ such that for all $M$ and all $Z,Z'$,
\[
\|F_M(Z)-F_M(Z')\| \le \kappa \|Z-Z'\|.
\]
\end{assumption}

By Theorems~\ref{thm:kl-eval-contraction} and~\ref{thm:parzen-improvement},
Assumption~\ref{as:uniform-contraction} holds and the ODE~\eqref{eq:fast-ode}
admits a unique globally asymptotically stable equilibrium
\[
Z_M^\star = (Q_M^\star,\mu_M^\star).
\]

\subsection{Perturbed Contraction and Fixed-Point Lipschitz Continuity}

We next quantify how the equilibrium depends on memory.

\begin{lemma}[Lipschitz Continuity of the Fixed Point]
\label{lem:fp-lipschitz}
Assume:
\begin{enumerate}
\item $F_M$ is a $\kappa$-contraction uniformly in $M$;
\item $F_M$ is Lipschitz in memory:
\[
\|F_{M'}(Z)-F_M(Z)\| \le L_F \|M'-M\| \quad \forall Z.
\]
\end{enumerate}
Then the fixed point satisfies
\[
\|Z_{M'}^\star - Z_M^\star\|
\;\le\;
\frac{L_F}{1-\kappa}\,\|M'-M\|.
\]
\end{lemma}

\begin{proof}
Using $Z_M^\star = F_M(Z_M^\star)$,
\begin{align*}
\|Z_{M'}^\star - Z_M^\star\|
&= \|F_{M'}(Z_{M'}^\star) - F_M(Z_M^\star)\| \\
&\le \|F_{M'}(Z_{M'}^\star) - F_{M'}(Z_M^\star)\|
     + \|F_{M'}(Z_M^\star) - F_M(Z_M^\star)\| \\
&\le \kappa \|Z_{M'}^\star - Z_M^\star\|
     + L_F \|M'-M\|.
\end{align*}
Rearranging yields the claim.
\end{proof}

\subsection{Slow Timescale: Memory Dynamics}

Memory updates take the stochastic approximation form
\begin{equation}
\label{eq:memory-update}
M_{t+1}
=
M_t + \rho_t \bigl[h(M_t,Z_t) + \xi^{(M)}_t\bigr],
\end{equation}
where $Z_t=(Q_t,\mu_t)$, $h$ is the expected update, and $\xi^{(M)}_t$ is a martingale
difference noise term.
The associated slow ODE is
\begin{equation}
\label{eq:memory-ode}
\frac{dM}{d\varsigma} = h(M),
\end{equation}
with $\varsigma$ denoting slow time.
By Assumption~\ref{as:twots-detailed}(iv), this ODE admits a compact attractor set
$ M_\infty$.

We also have $\|M_{t+1}-M_t\|=O(\rho_t)$, a.k.a., 
$\sup_t \|M_{t+1}-M_t\|/\rho_t < \infty$.
Under the distance~\eqref{def:memory-distance},
each of the following practical update schemes induces a change of order $\rho_t$:
\begin{enumerate}
\item \emph{Experience replay}: a new case is added with probability $\rho_t$,
      changing the induced base measure by $O(\rho_t)$.
\item \emph{Sliding window}: one case is replaced with probability $\rho_t$,
      yielding an $O(\rho_t)$ change.
\item \emph{Importance-weighted memory}: weights are updated by a step $\rho_t$
      and the induced base measure changes Lipschitzly.
\end{enumerate}
Hence $\|M_{t+1}-M_t\|=O(\rho_t)$ almost surely.

\subsection{Tracking Lemma for Two-Timescale SA}

We now state the tracking result used in the proof.

\begin{lemma}[Tracking Lemma]
\label{lem:tracking}
Suppose:
\begin{enumerate}
\item For each fixed $M$, the fast ODE~\eqref{eq:fast-ode} has a globally asymptotically
      stable equilibrium $Z_M^\star$;
\item $Z_M^\star$ is Lipschitz in $M$ as in Lemma~\ref{lem:fp-lipschitz};
\item The step sizes satisfy $\rho_t/\eta_t \to 0$.
\end{enumerate}
Then
\[
\|Z_t - Z_{M_t}^\star\| \;\to\; 0
\quad \text{almost surely}.
\]
\end{lemma}
Lemma~\ref{lem:tracking} follows from standard two-timescale stochastic approximation
results; see \cite{borkar2009stochastic,BorkarMeyn2000}.

\subsection{Derivation of the Tracking Recursion}

Define the tracking error
\[
\epsilon_t = \|Z_t - Z_{M_t}^\star\|.
\]
The fast update admits the form
\[
Z_{t+1}
=
Z_t + \eta_t\bigl(F_{M_{t+1}}(Z_t) - Z_t\bigr) + \eta_t \xi_t,
\]
with martingale noise $\xi_t$.
Using contraction of $F_M$, Lipschitz continuity of $Z_M^\star$, and the triangle
inequality, we obtain
\begin{equation}
\label{eq:tracking-recursion}
\epsilon_{t+1}
\;\le\;
(1-\eta_t\lambda)\epsilon_t
+ \eta_t L \|M_{t+1}-M_t\|
+ \eta_t \|\xi_t\|,
\end{equation}
where $\lambda=1-\kappa>0$ and $L$ is the constant from
Lemma~\ref{lem:fp-lipschitz}.
Since $\|M_{t+1}-M_t\|=O(\rho_t)$ and $\rho_t=o(\eta_t)$, the drift term is negligible
relative to the contraction.

\subsection{Convergence to the Limit Set}

By Lemma~\ref{lem:tracking}, $\epsilon_t\to 0$ almost surely.
Since $M_t\to M_\infty$, all limit points of $(Z_t,M_t)$ lie in
\[
\mathcal{L}
=
\{(Z_M^\star,M): M\in M_\infty\}.
\]
If $M_t\to M_\infty$, then $(Q_t,\mu_t)\to(Q_{M_\infty}^\star,\mu_{M_\infty}^\star)$
almost surely.

This completes the proof.

\section{Notations of MDP, SRDP, and Reflected MDP}

\label{ap:comparison}

Table~\ref{tab:mdp-srdp-remdp-notation} summarises and contrasts the notation used for the underlying Markov Decision Process (MDP), the Stateful Reflective Decision Process (SRDP) (Definition~\ref{def:llm-mmdp}), and its Markovian reformulation as the Reflected MDP (Definition~\ref{def:llm-induced-mdp-v2}). The underlying MDP models only the environment dynamics, abstracting away any internal structure of the agent. The SRDP extends this framework by explicitly modelling episodic memory and a two-stage decision mechanism, in which a retrieval action selects a memory case that conditions the LLM’s action generation. Since the evolving memory breaks the Markov property in the original state space, the Reflected MDP restores Markovianity by augmenting the state with memory and absorbing the fixed LLM behaviour into effective transition and reward kernels. This reformulation allows classical reinforcement learning analysis to be applied while preserving the full agent–memory–LLM interaction dynamics.

\begin{table}[t]
\centering
\small
\begin{tabular}{p{2.6cm} p{3.1cm} p{4.2cm} p{4.6cm}}
\toprule
\textbf{Concept} 
& \textbf{Underlying MDP} 
& \textbf{SRDP} 
& \textbf{Reflected MDP} \\
\midrule

Decision process
& $\mathcal D = \langle \mathcal S, \mathcal A, P, R, \gamma \rangle$
& $\mathcal D_{\mathrm{SRDP}} = \langle \mathcal S, \mathcal A, P, R, \gamma, \mathcal M, p_{\mathrm{LLM}} \rangle$
& $\mathcal D_{\mathrm{ReMDP}} = \langle \mathcal X, \mathcal C, P_{\mathrm{LLM}}, R_{\mathrm{LLM}}, \gamma \rangle$ \\

State space
& $s \in \mathcal S$
& $s \in \mathcal S$ (environment state)
& $x=(s,M)\in\mathcal X := \mathcal S \times \mathcal M$ \\

Action space
& $a \in \mathcal A$
& Two-stage: retrieval $c \in \mathcal C(M)$, then $a \in \mathcal A$
& Retrieval only: $c \in \mathcal C(M)$ \\

Memory
& None
& Episodic memory $M \in \mathcal M$, evolving via $\mathrm{Write}$
& Included in state: $M$ is part of $x=(s,M)$ \\

Retrieval action \newline set
& N/A
& $\mathcal C(M) := M$ (memory indexers)
& $\mathcal C(M) := M$ \\

Policy
& $\pi(a \mid s)$
& Composite policy:
$\displaystyle
\pi^\mu(a \mid s,M)
= \sum_{c \in M} \mu(c \mid s,M)\, p_{\mathrm{LLM}}(a \mid s,c)
$
& Retrieval policy $\mu(c \mid x)$ \\

LLM kernel
& N/A
& $p_{\mathrm{LLM}}(a \mid s,c)$
& Absorbed into $P_{\mathrm{LLM}}, R_{\mathrm{LLM}}$ \\

Transition kernel
& $P(s' \mid s,a)$
& $s' \sim P(\cdot \mid s,a)$, \\
& & $M'=\mathrm{Write}(M,s,a,r,s')$
& $P_{\mathrm{LLM}}(x' \mid x,c)
= \sum_a p_{\mathrm{LLM}}(a \mid s,c)\,
P(s' \mid s,a)$ \\

Reward
& $R(s,a)$
& $r=R(s,a)$
& $R_{\mathrm{LLM}}(x,c)
= \sum_a p_{\mathrm{LLM}}(a \mid s,c) R(s,a)$ \\

Markov property
& Markov in $s$
& Non-Markov in $s$ due to evolving $M$
& Markov in augmented state $x=(s,M)$ \\

Learning focus
& Action selection
& Retrieval + memory growth
& Retrieval policy optimisation \\

Role
& Models environment \newline dynamics
& Models agent--memory--LLM interaction
& Markovian control abstraction of SRDP \\

\bottomrule
\end{tabular}
\caption{Comparison of notations for the underlying MDP, Stateful Reflective Decision Process (SRDP), and the Reflected MDP.}
\label{tab:mdp-srdp-remdp-notation}
\end{table}

\end{document}